\documentclass{article}

\usepackage[utf8]{inputenc}
\usepackage{amsmath}
\usepackage{amssymb}
\usepackage{microtype}
\usepackage{url}
\usepackage{graphicx}

\usepackage{listings}
\usepackage{xcolor}
\usepackage{skak}

\providecommand{\sysfont}{\textit}

\newcommand{\aspviz}{\sysfont{aspviz}}

\newcommand{\clorm}{\sysfont{clorm}}
\newcommand{\clingraph}{\sysfont{clingraph}}

\newcommand{\Clingraph}{\sysfont{Clingraph}}

\newcommand{\asprilo}{\sysfont{asprilo}}

\newcommand{\clasp}{\sysfont{clasp}}

\newcommand{\clingo}{\sysfont{clingo}}

\newcommand{\dlv}{\sysfont{dlv}}

\newcommand{\graphviz}{\sysfont{graphviz}}

\newcommand{\python}{Python}

\providecommand{\Underscore}{\textunderscore}

\lstdefinelanguage{clingo}{basicstyle=\ttfamily,keywordstyle=[1]\bfseries,keywordstyle=[2]\bfseries,keywordstyle=[3]\bfseries,showstringspaces=false,literate={_}{\Underscore}1 {\%\%}{}0,escapeinside={\#(}{\#)},alsoletter={\#,\&},keywords=[1]{not,from,import,def,if,else,elif,return,while,break,and,or,for,in,del,and,class,with,as,is,yield,async},keywords=[2]{\#const,\#show,\#minimize,\#base,\#theory,\#count,\#external,\#program,\#script,\#end,\#heuristic,\#edge,\#project,\#show,\#sum},keywords=[3]{&,&dom,&sum,&diff,&show},morecomment=[l]{\#\ },morecomment=[l]{\%\ },morestring=[b]",stringstyle={\itshape},commentstyle={\color{darkgray}}}

\lstdefinelanguage{python}{basicstyle=\ttfamily,keywordstyle=[1]\bfseries,showstringspaces=false,literate={_}{\Underscore}{1},escapeinside={\#(}{\#)},alsoletter={\#,\&},keywords=[1]{not,from,import,def,if,else,elif,return,while,break,and,or,for,in,del,and,class,with,as,is,yield,async},morecomment=[l]{\#\ },morestring=[b]",stringstyle={\itshape},commentstyle={\color{darkgray}}}

 \lstdefinelanguage{clingos}{language=clingo,basicstyle=\small\ttfamily }

\usepackage[citestyle=numeric,natbib=true,backend=bibtex,maxbibnames=99]{biblatex}
\bibliography{krr,local,procs}

\begin{document}

\title{\Clingraph: A System for ASP-based Visualization\thanks{This paper is an extended version of an article presented at LPNMR'22~{\mbox{({Hahn et~al.~2022})}}.}} \date{}
\author{SUSANA HAHN
  \\
  \small{University of Potsdam, Germany, and Potassco Solutions}
  \and
  ORKUNT SABUNCU
  \\
  \small{TED University, Turkey, and Potassco Solutions}
  \and
  TORSTEN SCHAUB,
  TOBIAS STOLZMANN
  \\
  \small{University of Potsdam, Germany, and Potassco Solutions}}

\maketitle

\begin{abstract}
  We present the ASP-based visualization tool \clingraph,
  which aims at visualizing various concepts of ASP by means of ASP itself.
  This idea traces back to the \aspviz\ tool and \clingraph\
  redevelops and extends it in the context of modern ASP systems.
  More precisely, \clingraph\ takes graph specifications in terms of ASP facts and hands them over to the graph
  visualization system \graphviz.
  The use of ASP provides a great interface between logic programs and/or answer sets and their visualization.
  Also, \clingraph\ offers a \python\ API that extends this ease of interfacing to \clingo's API,
  and in turn to connect and monitor various aspects of the solving process.
\end{abstract}
 \section{Introduction}\label{sec:introduction}

With the advance of Answer Set Programming (ASP;~\cite{lifschitz19a}) into more and more complex application domains,
also the need for inspecting problems as well as their solution increases significantly.
The intrinsic difficulty lies in the fact that ASP constitutes a general problem solving paradigm,
whereas the wide spectrum of applications rather calls for customized presentations.

We address this by taking up the basic idea of \aspviz~\cite{clvobrpa08a},
to visualize ASP by means of ASP itself, and
extend it in the context of modern ASP systems.
The resulting system is called \clingraph\ (v1.1.0).\footnote{\url{https://github.com/potassco/clingraph}}$^,$\footnote{\url{https://clingraph.readthedocs.io}}
The common idea is to specify a visualization in terms of a logic program that defines special atoms capturing graphic
elements.
This allows us to customize the presentation of an application domain by means of ASP,
and thus to easily connect with the problem specification and its solutions.

The visualization in \clingraph\ rests upon graph structures that are passed on to the graph layout system \graphviz.\footnote{\url{https://graphviz.org}}
To this end,
\clingraph\ takes---in its basic setting---a set of facts over predicates
\lstinline{graph},
\lstinline{node},
\lstinline{edge}, and
\lstinline{attr}
as input, and produces an output visualizing the induced graph structure.

As a simple example,
consider the graph coloring problem in Listing~\ref{lst:color}.
\begin{lstlisting}[language=clingos,basicstyle=\small\ttfamily,label={lst:color},caption={Graph coloring instance, encoding and display (\texttt{color.lp})}]
node(1..6).%%#(\label{lst:color:instance:begin}#)
edge(1,2). edge(1,3). edge(1,4). edge(2,4). edge(2,5).
edge(2,6). edge(3,4). edge(3,5). edge(5,6).
color(red; green; blue).%%#(\label{lst:color:instance:end}#)

{ assign(N, C) : color(C) } = 1 :- node(N).%%#(\label{lst:color:encoding:begin}#)
:- edge(N, M), assign(N, C), assign(M, C).%%#(\label{lst:color:encoding:end}#)

#show node/1.%%#(\label{lst:color:show:node}#)
#show edge((N,M)) : edge(N, M).%%#(\label{lst:color:show:edge}#)
#show attr(graph_nodes, default, style, filled).%%#(\label{lst:color:show:attr:one}#)
#show attr(node, N, color, C) : assign(N, C).%%#(\label{lst:color:show:attr:two}#)
\end{lstlisting}
The actual problem instance and encoding are given in Lines~\ref{lst:color:instance:begin}--\ref{lst:color:instance:end}
and~\ref{lst:color:encoding:begin}--\ref{lst:color:encoding:end}, respectively.
However, of particular interest are Lines~\ref{lst:color:show:node}--\ref{lst:color:show:attr:two} that use
\lstinline{#show} directives to translate the resulting graph colorings into \clingraph's input format.
While Line~\ref{lst:color:show:node} and~\ref{lst:color:show:edge} account for the underlying graph,
the two remaining lines comprise instructions to \graphviz.
Line~\ref{lst:color:show:attr:one} fixes the layout of graph nodes.
More interestingly, Line~\ref{lst:color:show:attr:two} translates the obtained graph coloring to layout instructions for \graphviz.
Our omission of an atom over \lstinline{graph/1} groups all entities under a default graph labeled \lstinline{default}
(which can be changed via an option;
similarly, graphs are taken to be undirected unless changed by option \lstinline{--type}).

\begin{figure}[ht!]
  \centering
  \includegraphics[scale=0.25]{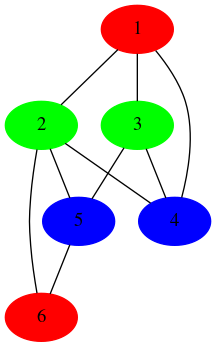}
  \caption{Visualization of the (first) stable model of the logic program in Listing~\ref{lst:color}}
  \label{fig:color}
\end{figure}

Launching \clingo\ so that only the resulting stable model is obtained as a set of facts allows us to visualize the result
via \clingraph:
\begin{lstlisting}[basicstyle=\small\ttfamily,numbers=none,xleftmargin=\parindent]
clingo --outf=0 -V0 --out-atomf=%s. color.lp | head -n1 | \
clingraph --out=render --format=png\end{lstlisting}
The used options suppress \clingo\ output and transform atoms into facts;
the intermediate UNIX command extracts the line comprising the stable model.
Note that one can also use a solver other than \clingo\ to generate the stable model in the expected form.
The final call to \clingraph\ produces a file in PNG format, shown in Figure~\ref{fig:color}.

Obviously, the above proceeding only reflects the very basic functionality of \clingraph.
We elaborate upon its extended functionality in the next section and
present a series of illustrative cases studies in Section~\ref{sec:cases}.
They range from the visualization of stable models,
including animated dynamic solutions as well as interactive ones,
over visualizing the actual solving process,
to the visual inspection of the structure of logic programs.
Finally, we present in Section~\ref{sec:template} some support for generating strings by using a template engine.
We summarize our approach and relate it to others' in Section~\ref{sec:discussion}.

 \section{\Clingraph}\label{sec:clingraph}

In its most basic setting, \clingraph\ can be regarded as a front-end to \graphviz\ that relies on the fact format
sketched above.
In fact, the full-fledged version of the fact format allows for specifying multiple graphs as well as subgraphs.
The former is done by supplying several instances of predicate \lstinline{graph/1} whose only argument provides an identifier
for regrouping all elements belonging to the graph at hand.
To that effect, there are also binary versions of predicates \lstinline{node} and \lstinline{edge},
whose second argument refers to the encompassing graph.
For example, the following facts describe \lstinline{n} graphs, each with one edge connecting two nodes.
\begin{lstlisting}[language=clingos]
id(1..n).%%#(\label{lst:multiple:ids}#)
graph(g(X)) :- id(X).%%#(\label{lst:multiple:graph}#)
node(n((a;b), X),g(X)) :- id(X).
edge((n(a,X),n(b,X)),g(X)) :- id(X).
\end{lstlisting}
Multiple graphs are of particular interest when visualizing dynamic domains, as in planning,
where each graph may represent a state of the world.
We illustrate this in Section~\ref{sec:cases} and show how the solution to a planning problem can be turned into an animation.

Subgraphs\footnote{Subgraphs correspond to clusters in \graphviz.}
are specified by the binary version of \lstinline{graph/2},
whose second argument indicates the super-ordinate graph.
For instance, replacing Line~\ref{lst:multiple:graph} above by the following two rules
makes \lstinline{g(X)} a subgraph of \lstinline{g(X+1)} for \lstinline{X=1..n-1}.
\begin{lstlisting}[numbers=none]
graph(g(X))        :- id(X), not id(X+1).
graph(g(X),g(X+1)) :- id(X),     id(X+1).
\end{lstlisting}

\Clingraph\ allows for selecting designated graphs by supplying their identifier to option \lstinline{--select-graph};
several ones are selected by repeating the option with the respective identifiers on the command line.

As mentioned, the quaternary predicate \lstinline{attr/4} describes properties of graph elements;
this includes all attributes of \graphviz.
The first argument fixes the type of the element, namely,
\lstinline{graph},
\lstinline{node}, and
\lstinline{edge}, along with keywords
\lstinline{graph_nodes} and
\lstinline{graph_edges}
to refer to all nodes and edges of a graph.
The second argument gives the identifier of the element,
and the last two provide the name and value of the \graphviz\ attribute.
Some attributes, mainly labels, are often constructed by concatenating multiple values.
We simplify this burden by providing an integration with a template engine to allow string formatting.
In Section~\ref{sec:template}, we describe this extension in detail.

In order to avoid name clashes,
\clingraph\ offers the option \lstinline{--prefix} to change all graph-oriented predicates by prepending a common prefix.
For instance, \lstinline{--prefix='viz-'} changes the dedicated predicate names to \lstinline{viz-graph},
\lstinline{viz-node}, \lstinline{viz-edge}, and \lstinline{viz-attr} while maintaining their arities.

The more interesting use cases emerge by using \emph{visualization encodings}.
While in our introductory example, the latter was mimicked by \lstinline{#show} statements,
in general, a visualization encoding can be an arbitrary logic program producing atoms over the four graph-oriented predicates.
Obviously, when it comes to visualization, a given problem encoding can then be supplemented with a dedicated
visualization encoding, whose output is then visualized by \clingraph\ as shown in the introductory section.

In practice, however, it turns out that this joint approach often results in a significant deceleration of the solving process.
Rather, it is often advantageous to resort to a sequential approach, in which the stable models of the problem encoding
are passed to a visualization encoding.
This use case is supported by \clingraph\ with extra functionality when using the ASP system \clingo.
More precisely, this functionality relies upon the \clingo\ feature to combine the output of a run, possibly comprising
various stable models, in a single \lstinline{json} object.\footnote{\url{https://www.json.org}}
To this end, \clingraph\ offers the option \lstinline{--select-model} to select one or multiple stable models from the
\lstinline{json} object.
Multiple models are selected by repeating the option with the respective number.

To illustrate this, let us replace Line~\ref{lst:multiple:ids} above by
\begin{lstlisting}[numbers=none]
{ id(1..n) } = 1.
\end{lstlisting}
to produce \lstinline{n} stable models with one graph each, rather than a single model with \lstinline{n} graphs as above.
The handover of all stable models of the resulting logic program in \lstinline{multiple.lp} to \clingraph\
can then be done by the following command:
\begin{lstlisting}[basicstyle=\small\ttfamily,numbers=none,xleftmargin=\parindent]
clingo --outf=2 -c n=10 0 multiple.lp | \
clingraph --out=tex --select-model=0 --select-model=9
\end{lstlisting}
The option \lstinline{--outf=2} instructs \clingo\ to produce a single \lstinline{json} object as output.
We request all \lstinline{10} stable models via `\lstinline{-c n=10 0}'.
Then, \clingraph\ produces a \LaTeX\ file depicting the graphs described in the first and tenth stable model.

In the quite frequent case that the stable models are produced exclusively by the problem encoding,
an explicit visualization encoding can be supplied via option \lstinline{--viz-encoding} to make \clingraph\ internally
produce the graphic representation from the given stable models
employing the \clingo\ API.
To ease the development of visualization encodings,
\clingraph\ also provides a set of external \python\ functions
(see Section~\ref{sec:cases} for an example).

Just like \clingraph's input, also its output may consist of one or several graph representations.
The specific representation is controlled by option \lstinline{--out} that can take the following values:
\begin{itemize}
\item \lstinline{facts}   \ produces the facts obtained after preprocessing (default)
\item \lstinline{dot}     \ produces graph representations in the language DOT
\item \lstinline{render}  \ generates images with the rendering method of \graphviz\
\item \lstinline{animate} \ generates a GIF after rendering
\item \lstinline{tex}     \ produces a \LaTeX\ file
\end{itemize}
The default option \lstinline{facts} allows us to inspect the processed input to \clingraph\ in fact format.
This involves the elimination of atoms irrelevant to \clingraph\ as well as the normalization of the graph representation
(e.g.,\ turning unary predicates \lstinline{node} and \lstinline{edge} into binary ones, etc.).
Options \lstinline{dot} and \lstinline{tex} result in text-based descriptions of graphs in the languages DOT and \LaTeX.
These formats allows for further post-processing and editing upon document integration.
The \LaTeX\ file is produced with \sysfont{dot2tex}.\footnote{\url{https://dot2tex.readthedocs.io}}
Arguments to \sysfont{dot2tex} can be passed through \clingraph\ via \lstinline{--tex-param}.
At long last, the options \lstinline{render} and \lstinline{animate} synthesize images for the graphs at hand.
While the former aims at generating one image per graph, the latter allows us to combine several graphs in an animation.
The format of a rendered graph is determined by option \lstinline{--format};
it defaults to PDF and alternative formats include PNG and SVG (cf.~Section~\ref{subsec:svg}).
Animation results in a  GIF file.
It is supported by options \lstinline{--fps} to fix the number of frames per second and \lstinline{--sort} to fix the
order of the graphs' images in the resulting animation.
The latter provides a handful of alternatives to describe the order in terms of the graph identifiers.

Also, it is worth mentioning that \clingraph's option \lstinline{--engine} allows us to choose among the eight layout
engines of \graphviz;\footnote{\url{http://www.graphviz.org/docs/layouts}} it defaults to \lstinline{dot} which is optimized for drawing directed graphs.

Last but not least,
\clingraph\ also offers an application programming interface (API) for \python.
Besides \graphviz, it heavily relies on \clorm,\footnote{\url{https://github.com/potassco/clorm}}
a \python\ library providing an Object Relational Mapping (ORM) interface to \clingo.
Accordingly, the major components of \clingraph's API are
its \texttt{Factbase} class, providing functionality for manipulating sets of facts via \clorm, and
the \texttt{graphviz} package, gathering functionality for interfacing to \graphviz.
We refer the interested reader to the API documentation for further details.\footnote{\url{https://clingraph.readthedocs.io/en/latest/clingraph/api.html}}
In conjunction with \clingo,
the API can be used for visualizing the solving process.
Two natural interfaces for this are provided by the \lstinline{on_model} callback of \clingo's \lstinline{solve} method as
well \clingo's \lstinline{Propagator} class.
For example, the former would allow for visualizing the intermediate stable models obtained when converging to an optimal
model during optimization.
The latter provides an even more fine-grained approach that allows for monitoring the search process by visualizing partial assignments (cf.~Section~\ref{subsec:vizsolving}).
 \section{Case studies}\label{sec:cases}

In this section, we list several case studies as examples that showcase various features of \clingraph.
Our first example, in Section~\ref{subsec:queens}, visualizes a solution of the well-known Queens puzzle.
The next one, in Section~\ref{subsec:robots}, aims at visualizing a dynamic problem of a robotic intra-logistics scenario where
the resulting animation points out temporal aspects of the problem.
Then, in Section~\ref{subsec:svg}, we explore interactivity in visualizations via \clingraph's SVG extension.
Finally, the last two case studies concentrate on visualizing aspects other than solutions of a problem.
To that end, these approaches visualize
the solving process of the solver (Section~\ref{subsec:vizsolving}) and
the structure of the input program (Section~\ref{subses:vizprogram}).

Many of these case studies need complex attributes, mainly labels,
which are composed of various values.
Thanks to the template engine integrated in \clingraph,
one can specify such an attribute conveniently by a template string having variables.
Then, separate rules may set values of these variables in a modular way.
Note that details on the usage of templates can be found in Section~\ref{sec:template}.

The interested reader is referred for further details on these examples and many others to \clingraph's distribution.\footnote{\url{https://github.com/potassco/clingraph/tree/master/examples}}

\subsection{Visualizing a solution of the Queens puzzle}\label{subsec:queens}

\begin{figure}[ht!]
  \centering
  \includegraphics[scale=0.25]{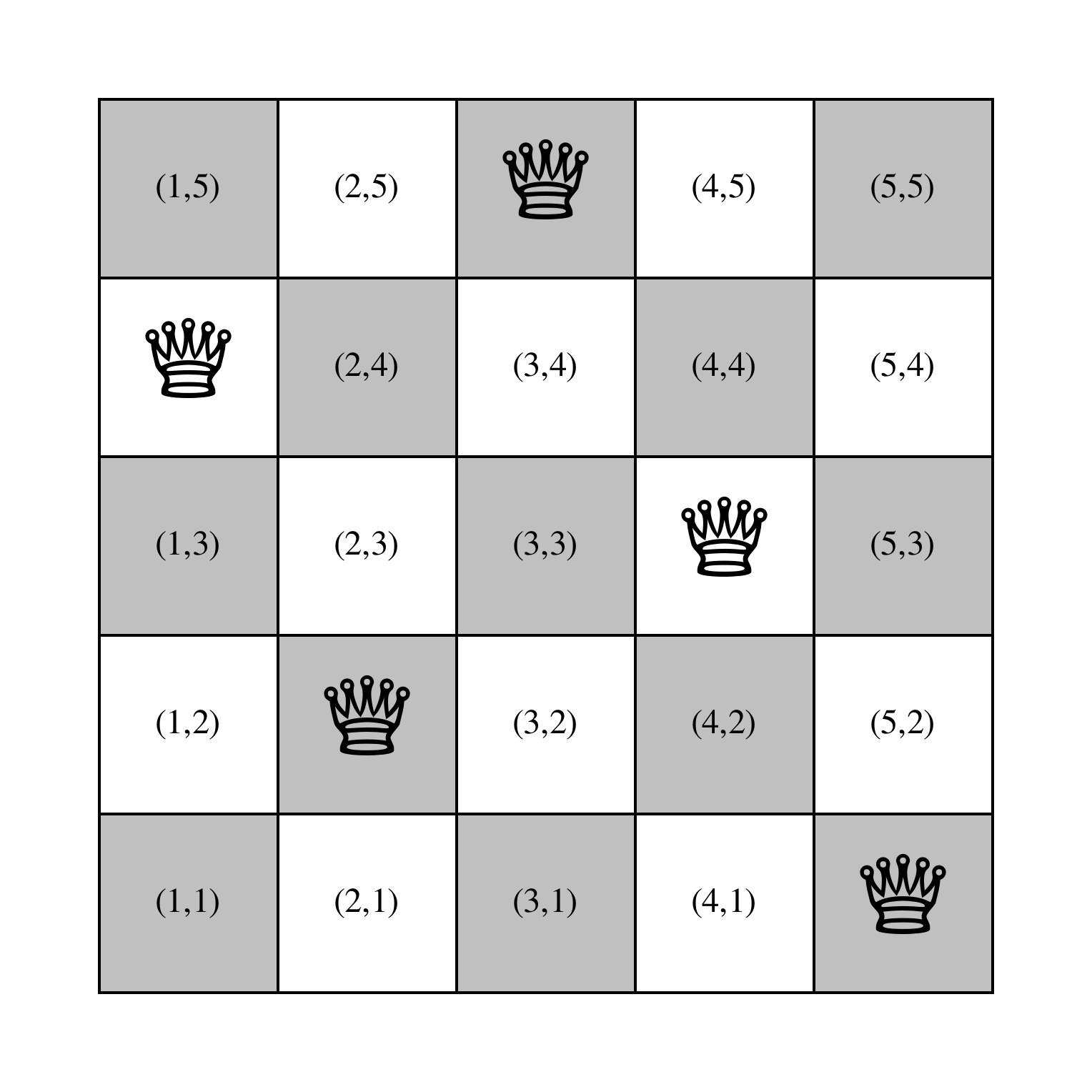}
  \caption{Visualization of (first) stable model of the logic program in Listing~\ref{lst:queens}}
  \label{fig:queens}
\end{figure}

As a first example,
consider the encoding of the Queens puzzle in Listing~\ref{lst:queens}.\footnote{\url{https://github.com/potassco/clingraph/tree/master/examples/queens}}
The idea is to place $n$ queens on an $n\times n$ chessboard so that no two queens attack one another.
\begin{lstlisting}[float,language=clingo,label={lst:queens},basicstyle=\small\ttfamily,caption={Queens puzzle (\texttt{queens.lp})}]
1 { queen(I,1..n) } 1 :- I = 1..n.
1 { queen(1..n,J) } 1 :- J = 1..n.
 :- 2 { queen(D-J,J) }, D = 2..2*n.
 :- 2 { queen(D+J,J) }, D = 1-n..n-1.

cell(1..n,1..n).
\end{lstlisting}
A solution is captured by atoms over predicate \lstinline{queen/2}.
The one comprised in the first stable model of \lstinline{queens.lp} for \lstinline{n=5}
is depicted in Figure~\ref{fig:queens}.
First of all, we note that the actual graph is laid out as a $5\times 5$ grid of white and gray squares.
Each atom \lstinline[mathescape]{queen($x$,$y$)} is then represented by putting the symbol \symqueen\ on the square with
coordinate $(x,y)$.
All other squares are simply labeled with their actual coordinate.

The visualization encoding producing the chessboard in Figure~\ref{fig:queens} is given in Listing~\ref{lst:queens:viz};
\begin{lstlisting}[float,language=clingo,label={lst:queens:viz},basicstyle=\small\ttfamily,caption={Visualization encoding for Queens puzzle (\texttt{viz.lp})}]
node((X,Y)) :- cell(X,Y).%%#(\label{lst:queens:node}#)

attr(node,(X,Y),width,1) :- cell(X,Y).%%#(\label{lst:queens:width}#)
attr(node,(X,Y),shape,square) :- cell(X,Y).%%#(\label{lst:queens:shape}#)
attr(node,(X,Y),style,filled) :- cell(X,Y).%%#(\label{lst:queens:style}#)
attr(node,(X,Y),fillcolor,gray) :- cell(X,Y),(X+Y)\2 = 0.%%#(\label{lst:queens:fillcolor:gray}#)
attr(node,(X,Y),fillcolor,white) :- cell(X,Y),(X+Y)\2 != 0.%%#(\label{lst:queens:fillcolor:white}#)
attr(node,(X,Y),fontsize,"50") :- queen(X,Y).%%#(\label{lst:queens:fontsize}#)
attr(node,(X,Y),label,"%%#(\symqueen#)") :- queen(X,Y).%%#(\label{lst:queens:label}#)
attr(node,(X,Y),pos,@pos(X,Y)) :- cell(X,Y).%%#(\label{lst:queens:pos}#)
\end{lstlisting}
it is used to generate the PDF in Figure~\ref{fig:queens} in the following way.
\begin{lstlisting}[basicstyle=\small\ttfamily,numbers=none,xleftmargin=\parindent]
clingo queens.lp -c n=5 --outf=2 | \
clingraph --viz-encoding=viz.lp --out=render --engine=neato
\end{lstlisting}
To better understand the visualization encoding,
it is important to realize that we use \sysfont{neato} as layout engine,
since it is better-suited for dealing with coordinates than the default engine \sysfont{dot}.

Let us now have a closer look at the encoding in Listing~\ref{lst:queens:viz}.
Interestingly, our graph consists of nodes only; no edges are provided.
This is because nodes are explicitly positioned and no edges are needed to connect them.
More precisely, one node is introduced in Line~\ref{lst:queens:node} for each cell of the chessboard.\footnote{Strictly speaking, the definition of predicate \lstinline{cell/2} belongs to the visualization encoding.
  Nonetheless, we add it to the problem encoding since the dimension of the board, viz.\ \lstinline{n}, is unavailable
  in the visualization encoding.
  This is a drawback of the sequential approach: information must be shared via the stable models.}
The remainder of the encoding is concerned with the layout and positioning of each individual node,
as reflected by the first and second argument of all remaining atoms over \lstinline{attr/4}.
This is done in a straightforward way in Lines~\ref{lst:queens:width} to~\ref{lst:queens:style} to fix the
\lstinline{width}, \lstinline{shape}, and \lstinline{style} of each node.
Line~\ref{lst:queens:fillcolor:gray} and~\ref{lst:queens:fillcolor:white} care about the alternating coloration of
nodes, depending on whether the sum of their coordinates is even or odd.
The next two lines deal with cells occupied by queens.
Unlike the previous rules that only refer to the problem instance,
here the derived attributes depend on the obtained solution.
That is,
for each atom \lstinline[mathescape]{queen($x$,$y$)},
Line~\ref{lst:queens:fontsize} fixes the \lstinline{fontsize}
of the \lstinline{label} \symqueen\
attributed to node \lstinline[mathescape]{($x$,$y$)} in Line~\ref{lst:queens:label}.
Whenever no \lstinline{label} is given to a node, its name is used instead,
as witnessed by Figure~\ref{fig:queens}.
Finally, Line~\ref{lst:queens:pos} handles the positioning of nodes.
In \sysfont{neato}, positions are formatted by two comma-separated numbers and entered
in a node’s \lstinline{pos} attribute.
If an exclamation mark `\lstinline{!}' is given as a suffix, the node is also pinned down.
The necessary transformation from pairs of terms is implemented by the external \python\ function \lstinline{pos(x,y)}
provided by \clingraph.
This function turns a node identifier \lstinline[mathescape]{($x$,$y$)} into a string of form \lstinline[mathescape]{"$x$,$y$!"}.
For each node, the result is then inserted as the fourth argument of predicate \lstinline{attr/4} in Line~\ref{lst:queens:pos}.
 \subsection{Visualizing dynamic problems}\label{subsec:robots}

As a second example,
let us look at a dynamic problem whose solutions can be visualized in terms of animations.
To this end, we have chosen a robotic intra-logistics scenario from the \asprilo\ framework~\cite{geobotscsangso18a}.
This scenario amounts to an extended multi-agent pathfinding problem having robots transport shelves to picking
stations and back somewhere.
The goal is to satisfy a batch of orders by transporting shelves covering all requested products to the picking station.
For brevity, we do not reproduce the actual problem encoding\footnote{\url{https://github.com/potassco/asprilo-encodings}} here and
rather restrict our attention to the input to the visualization encoding.
The input consists of action and fluent atoms accounting for a solution and how it progresses the problem scenario over time,
namely,
\begin{itemize}
\item \lstinline[mathescape]{move(robot($r$),($d_x$,$d_y$),$t$)}\footnote{We refrain from visualizing pickup and putdown actions, and rather represent them implicitly.}
  and
\item \lstinline[mathescape]{position($o$,($x$,$y$),$t$)} for $o$ among
  \lstinline[mathescape]{robot($r$)},
  \lstinline[mathescape]{shelf($s$)}, and
  \lstinline[mathescape]{station($p$)}.
\end{itemize}
A \lstinline[mathescape]{move} atom indicates that a robot $r$ moves in the cardinal direction
\lstinline[mathescape]{($d_x$,$d_y$)} at time step~$t$
(for $d_x,d_y\in\{-1,0,1\}$ such that $|d_x+d_y|=1$).
A \lstinline[mathescape]{position} atom tells us that object~$o$ is at position \lstinline[mathescape]{($x$,$y$)} at time step $t$.
All atoms sharing a common time step capture a state induced by the resulting plan.

The idea of the visualization encoding is now to depict a sequence of such states by combining the visualizations of
individual states in an animation.
Each state is represented by a graph that lays out the grid structure of a warehouse.
We use consecutive time steps to identify and to order these graphs.
This results in an atom \lstinline[mathescape]{graph($t$)} for each time step $t$.
Similarly, we identify nodes with their coordinate along with a timestamp.
This is necessary because nodes require a unique identifier across all (sub)graphs.
As well, we use edges indexed by time steps to trace (the last) movements.
\begin{itemize}
\item \lstinline[mathescape]{node((($x$,$y$),$t$),$t$)}
\item \lstinline[mathescape]{edge(((($x'$,$y'$),$t$),(($x'+d_x$,$y'+d_y$),$t$)),$t$)}
\end{itemize}
The first atom expresses that node \lstinline[mathescape]{(($x$,$y$),$t$)} belongs to graph $t$.
Similarly, the second one tells us that the edge
from node \lstinline[mathescape]{(($x'$,$y'$),$t$)}
to node \lstinline[mathescape]{(($x'+d_x$,$y'+d_y$),$t$)}
belongs to graph $t$.
It is induced by an action \lstinline[mathescape]{move(robot($r$),($d_x$,$d_y$),$t$)} and its precondition
\lstinline[mathescape]{position(robot($r$),($x'$,$y'$),$t-1$)}.

\begin{figure}[!ht]
  \centering
  \frame{\includegraphics[scale=0.14]{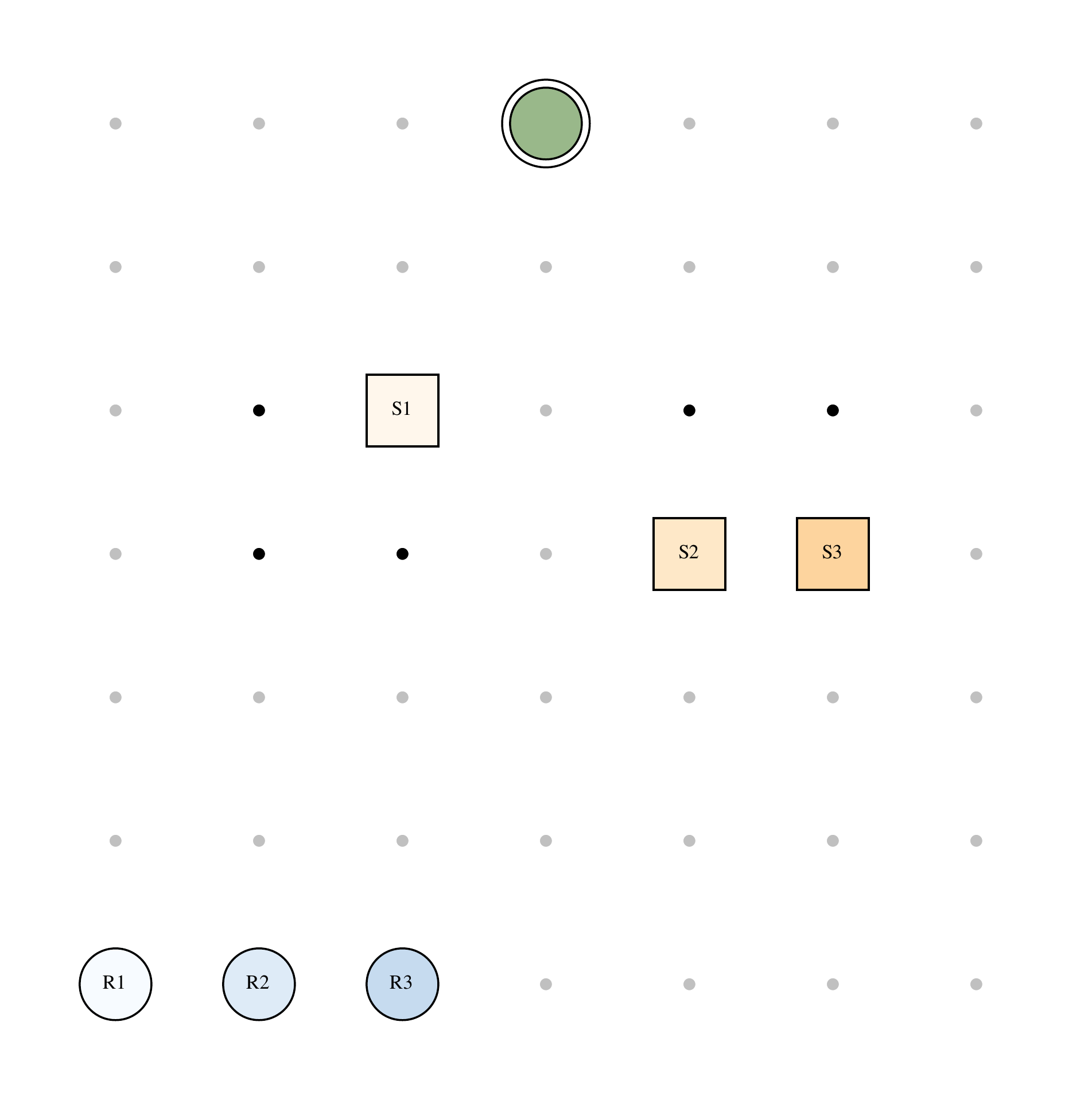}}
  \frame{\includegraphics[scale=0.14]{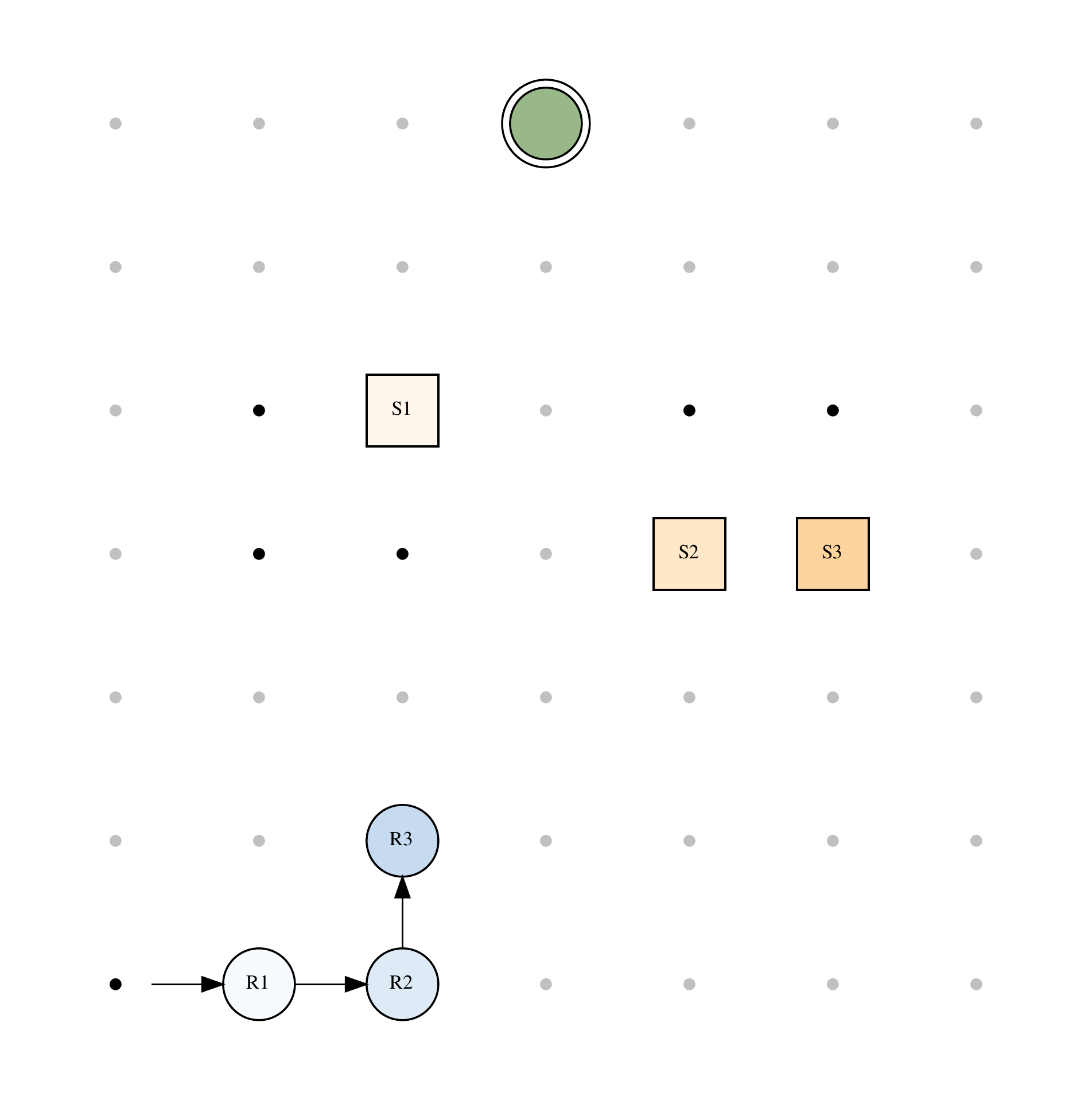}}
  \frame{\includegraphics[scale=0.14]{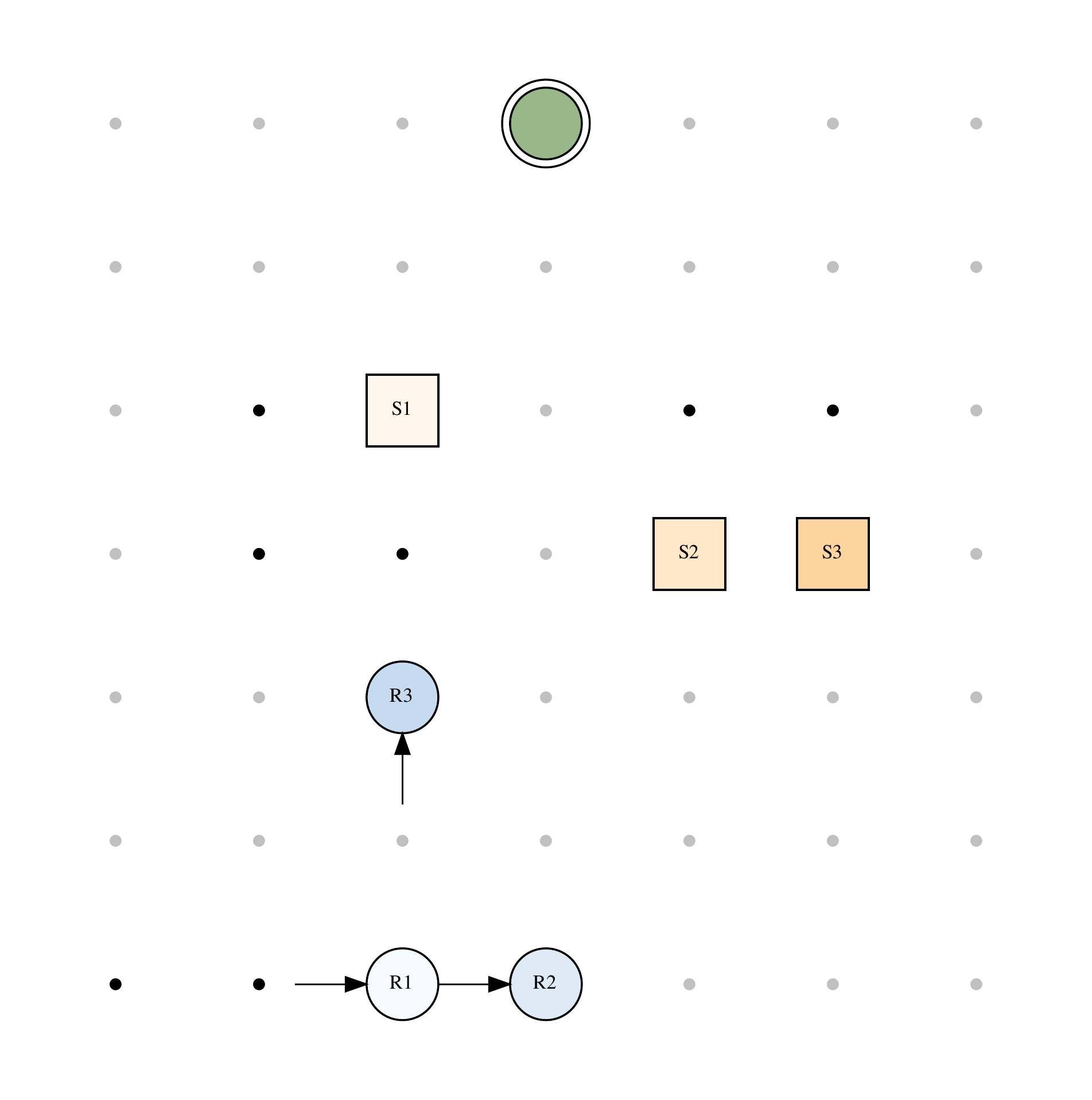}}
  \frame{\includegraphics[scale=0.14]{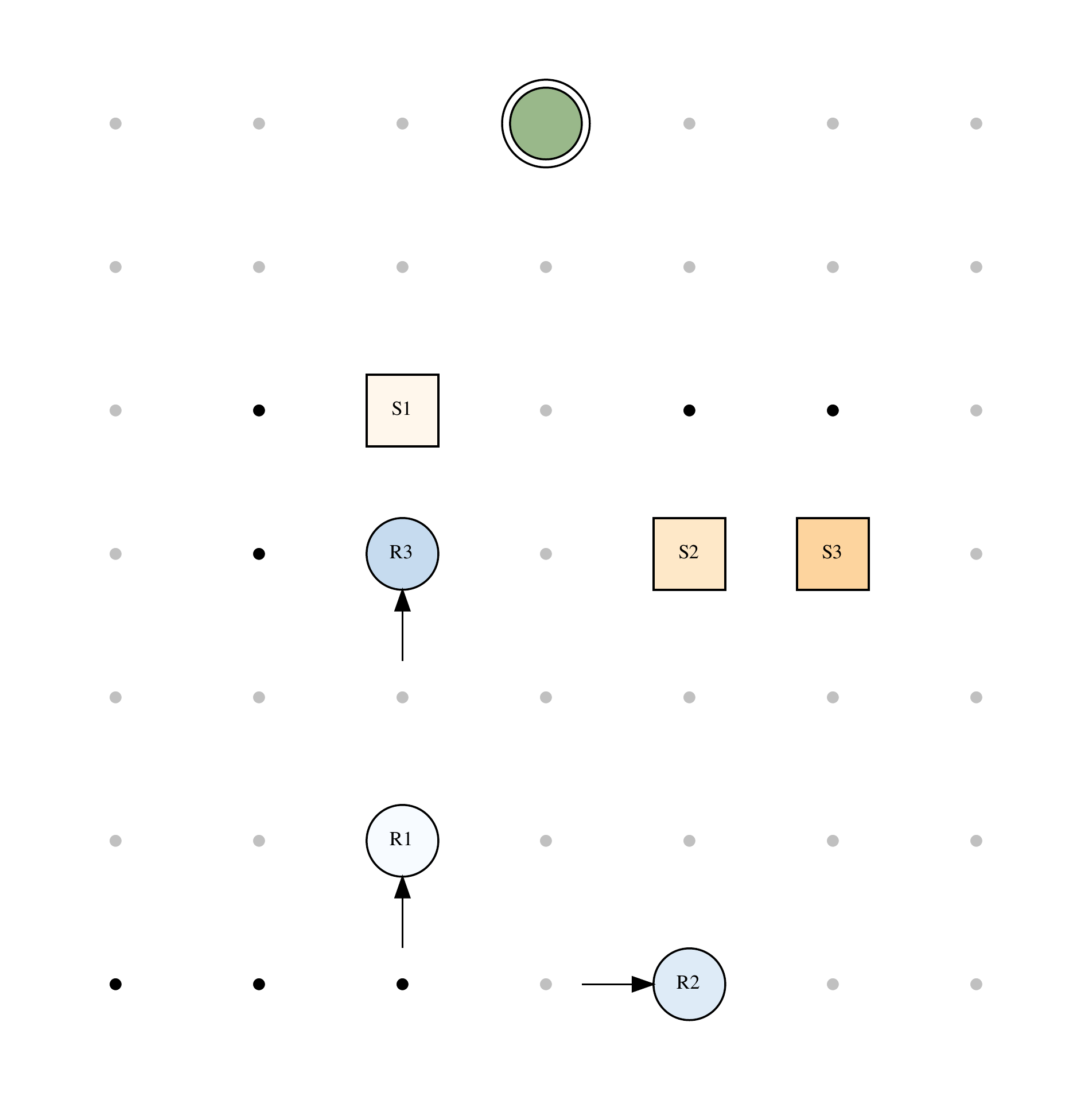}}
  \frame{\includegraphics[scale=0.14]{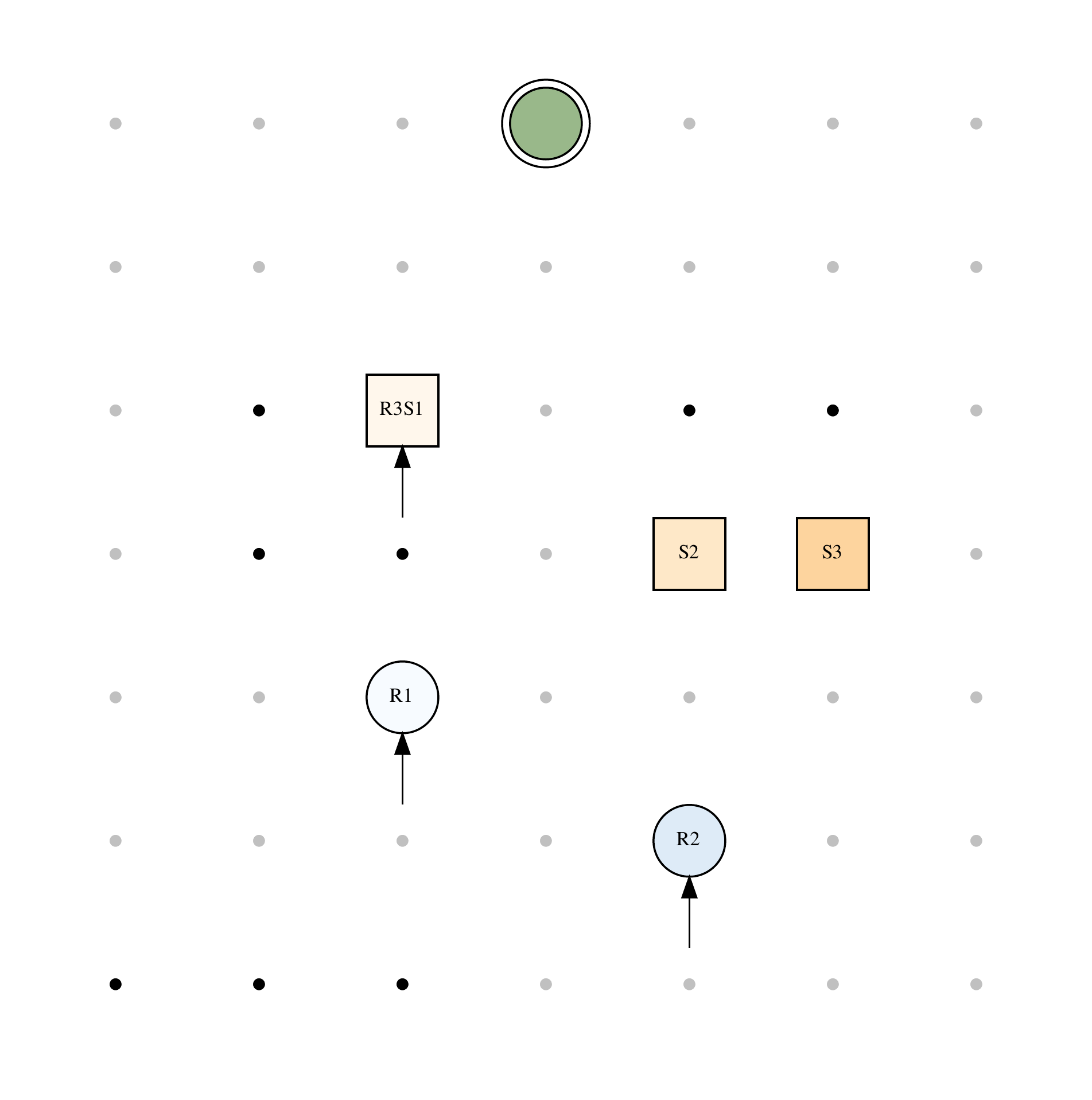}}
  \frame{\includegraphics[scale=0.14]{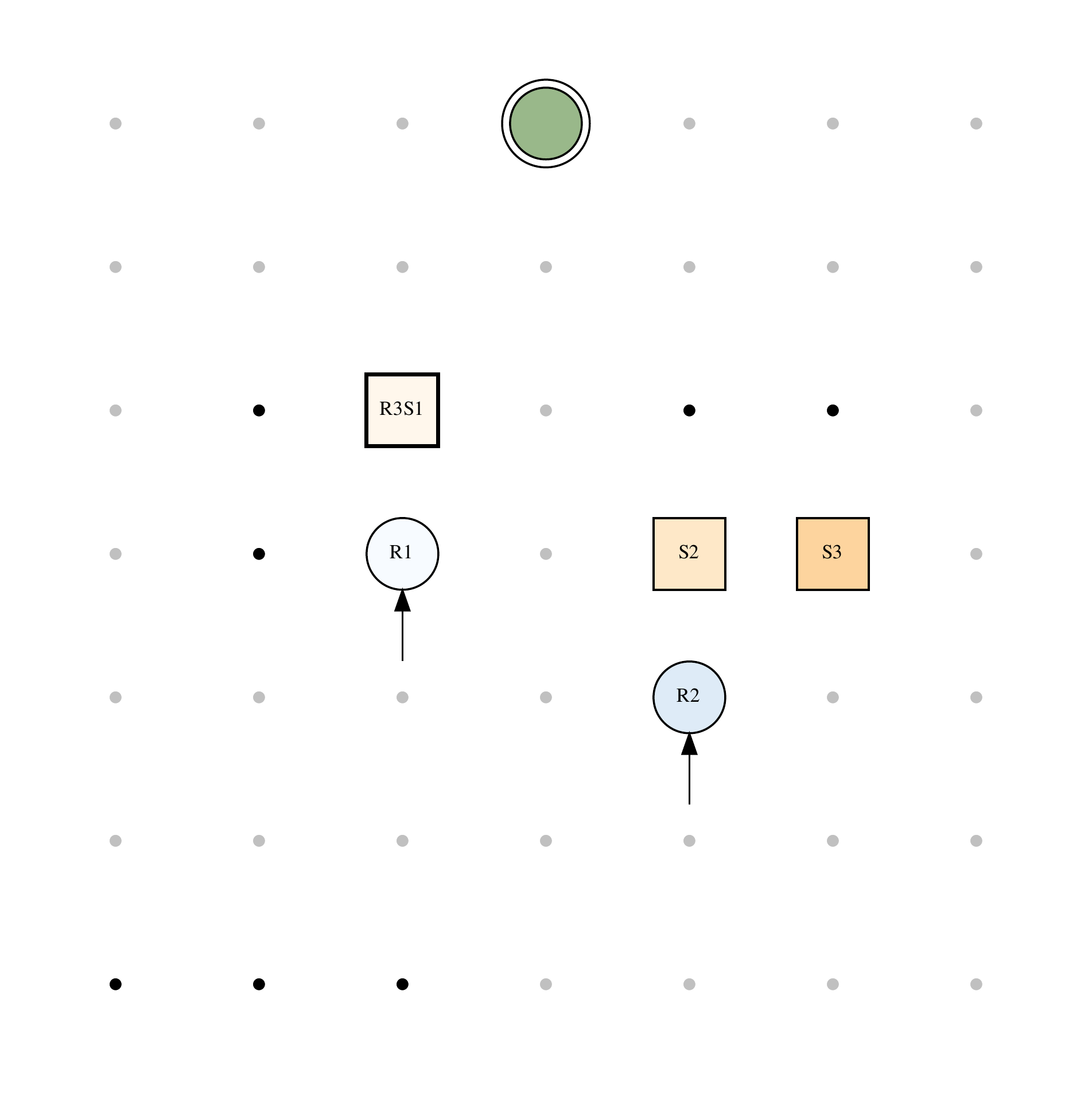}}
  \frame{\includegraphics[scale=0.14]{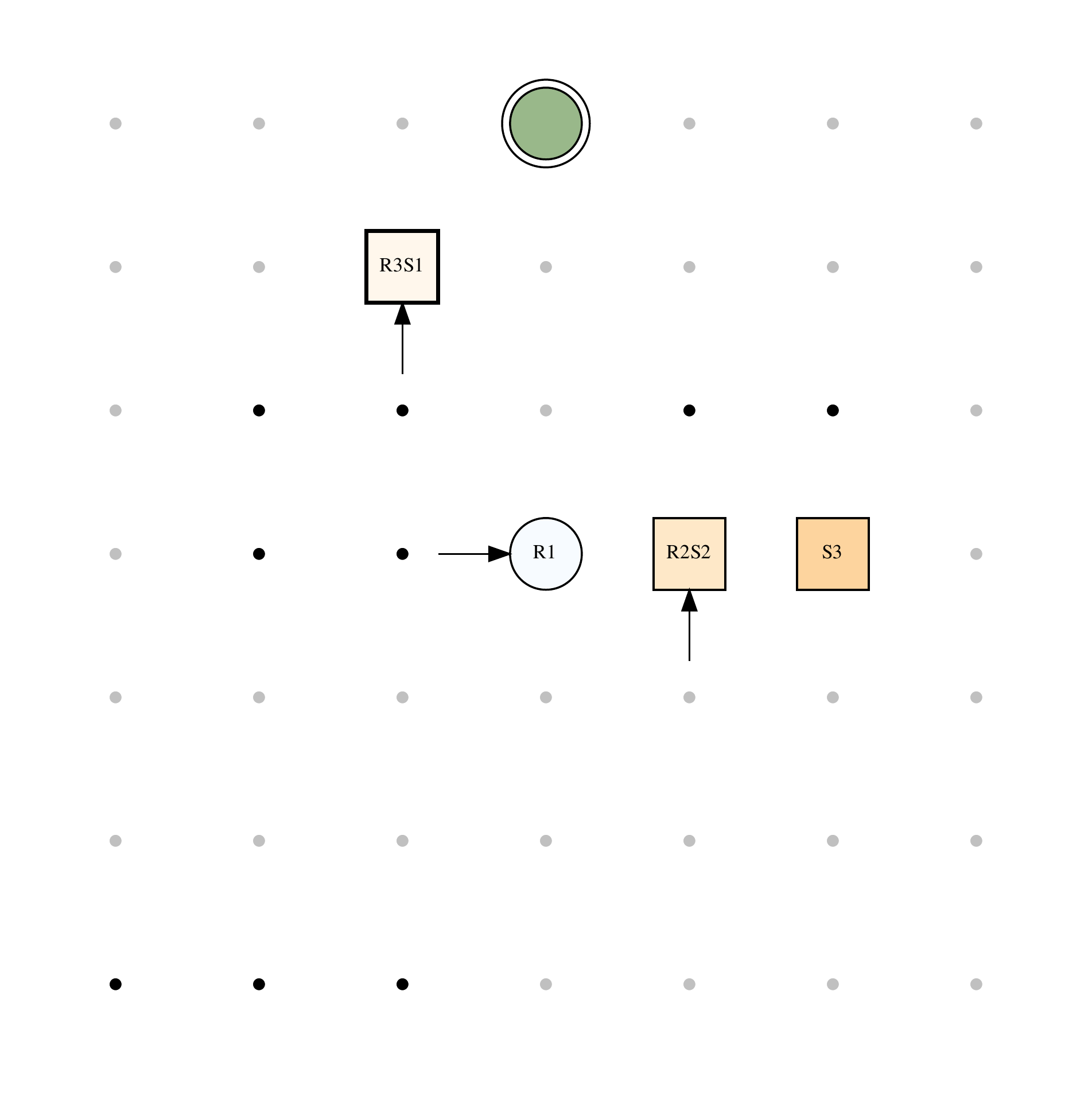}}
  \frame{\includegraphics[scale=0.14]{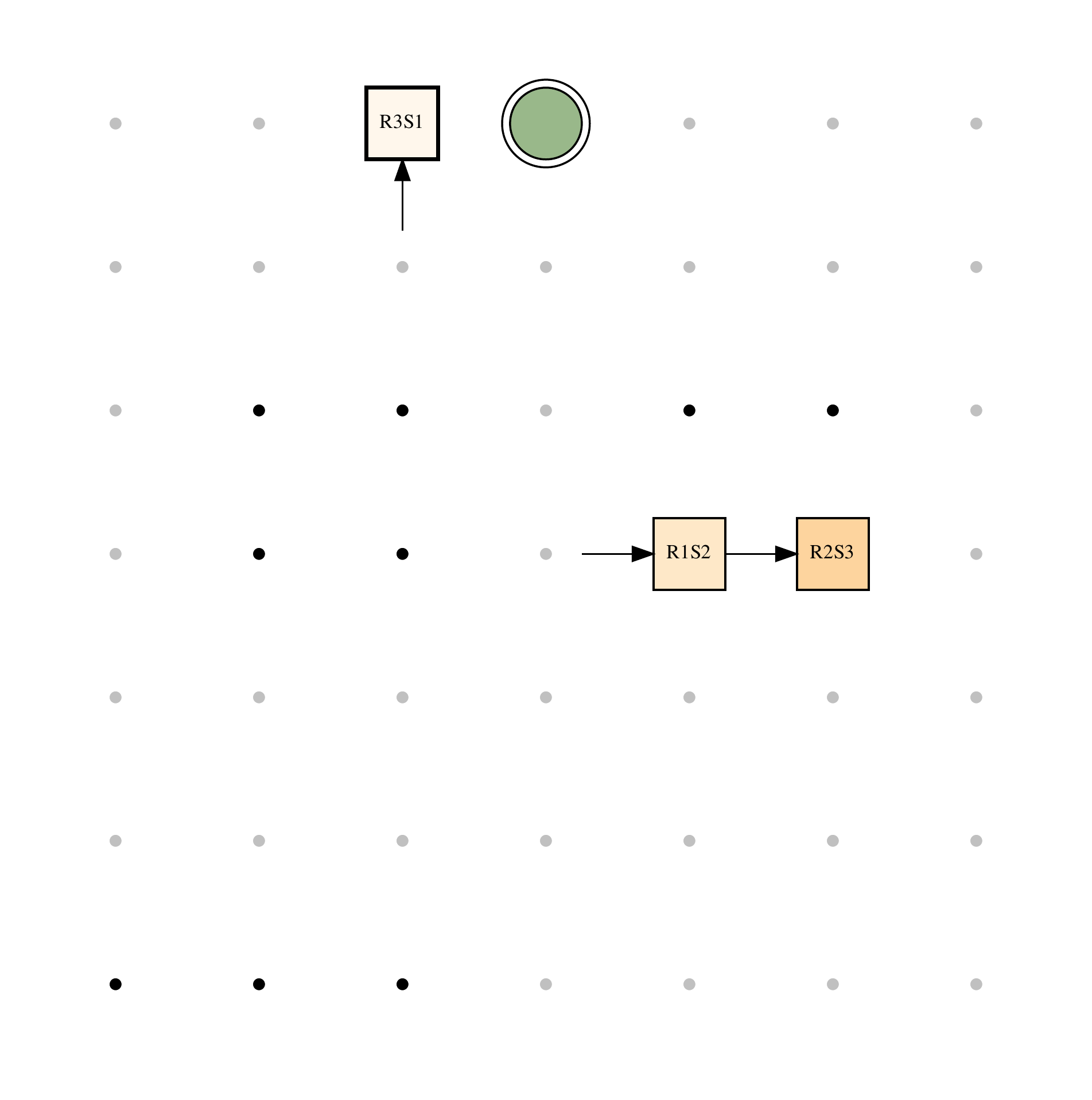}}
  \frame{\includegraphics[scale=0.14]{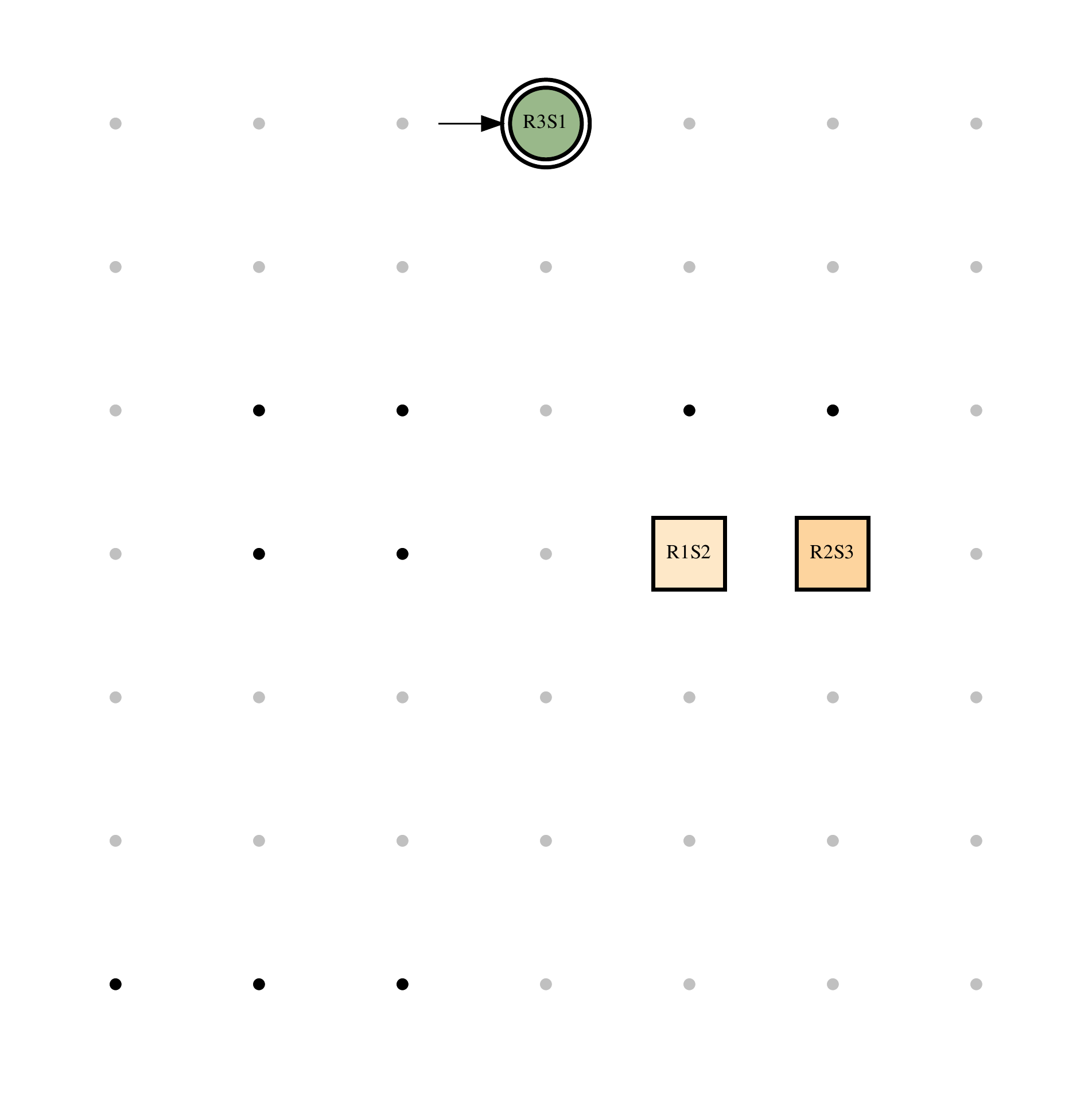}}
  \frame{\includegraphics[scale=0.14]{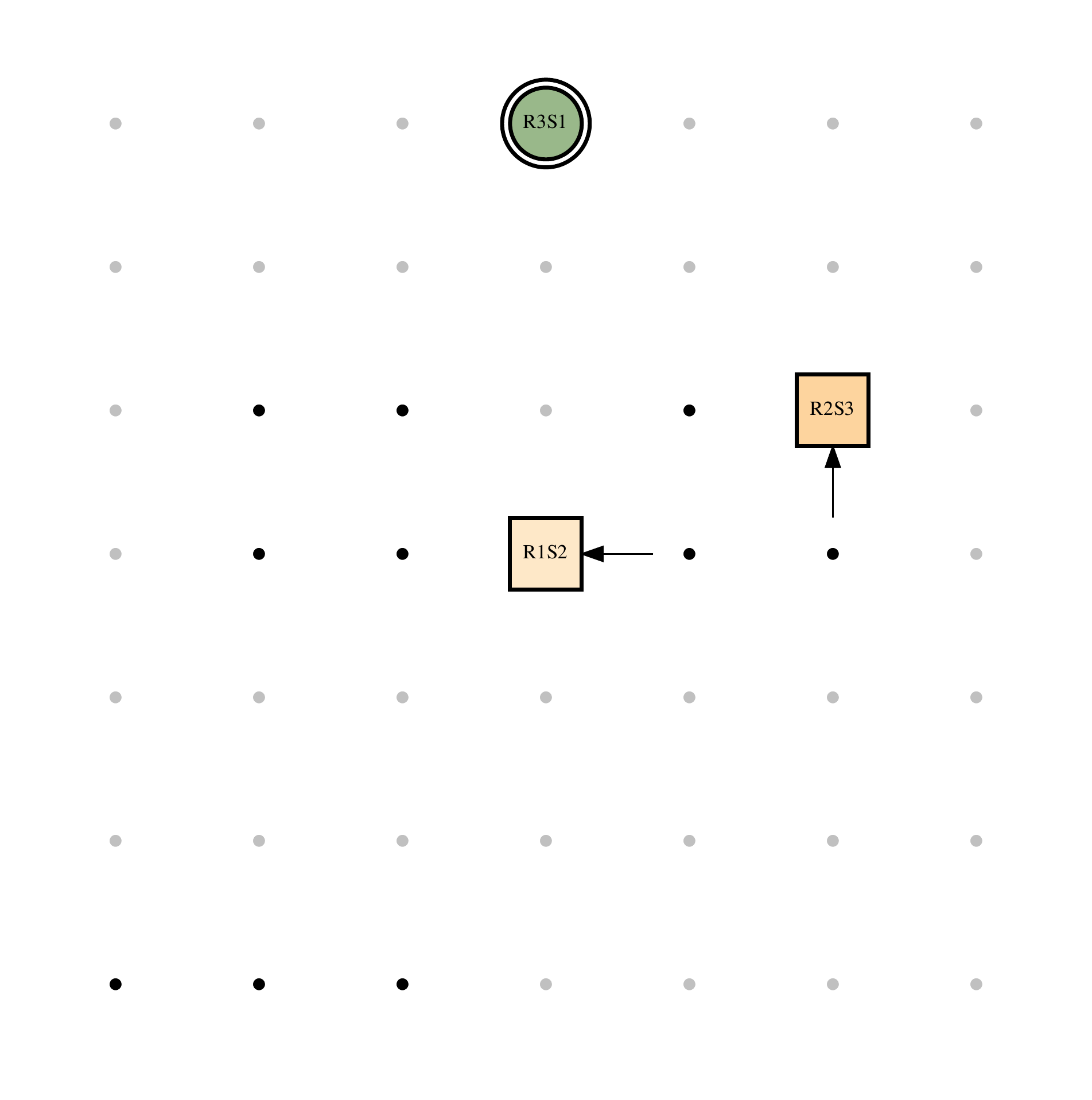}}
  \frame{\includegraphics[scale=0.14]{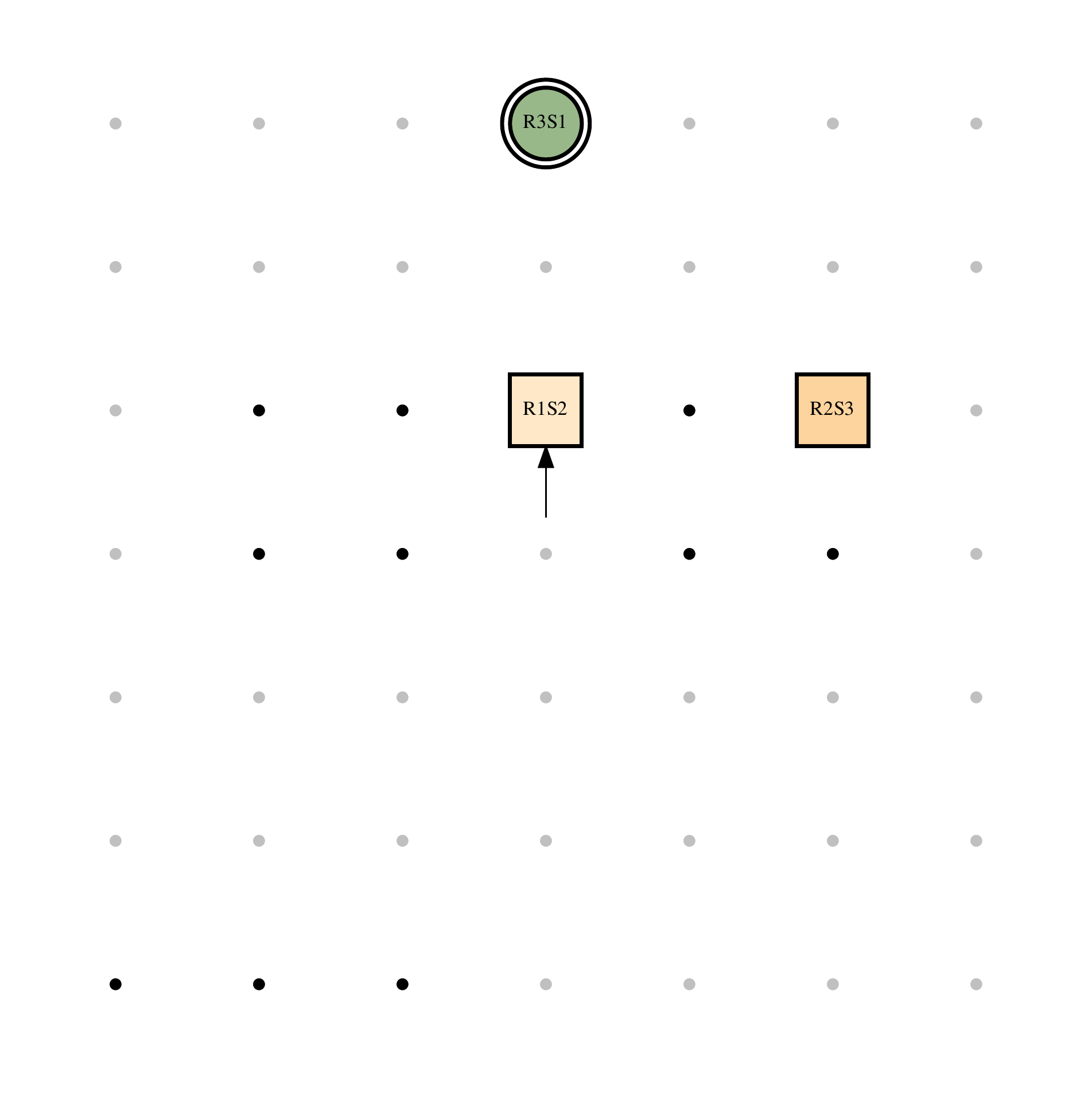}}
  \frame{\includegraphics[scale=0.14]{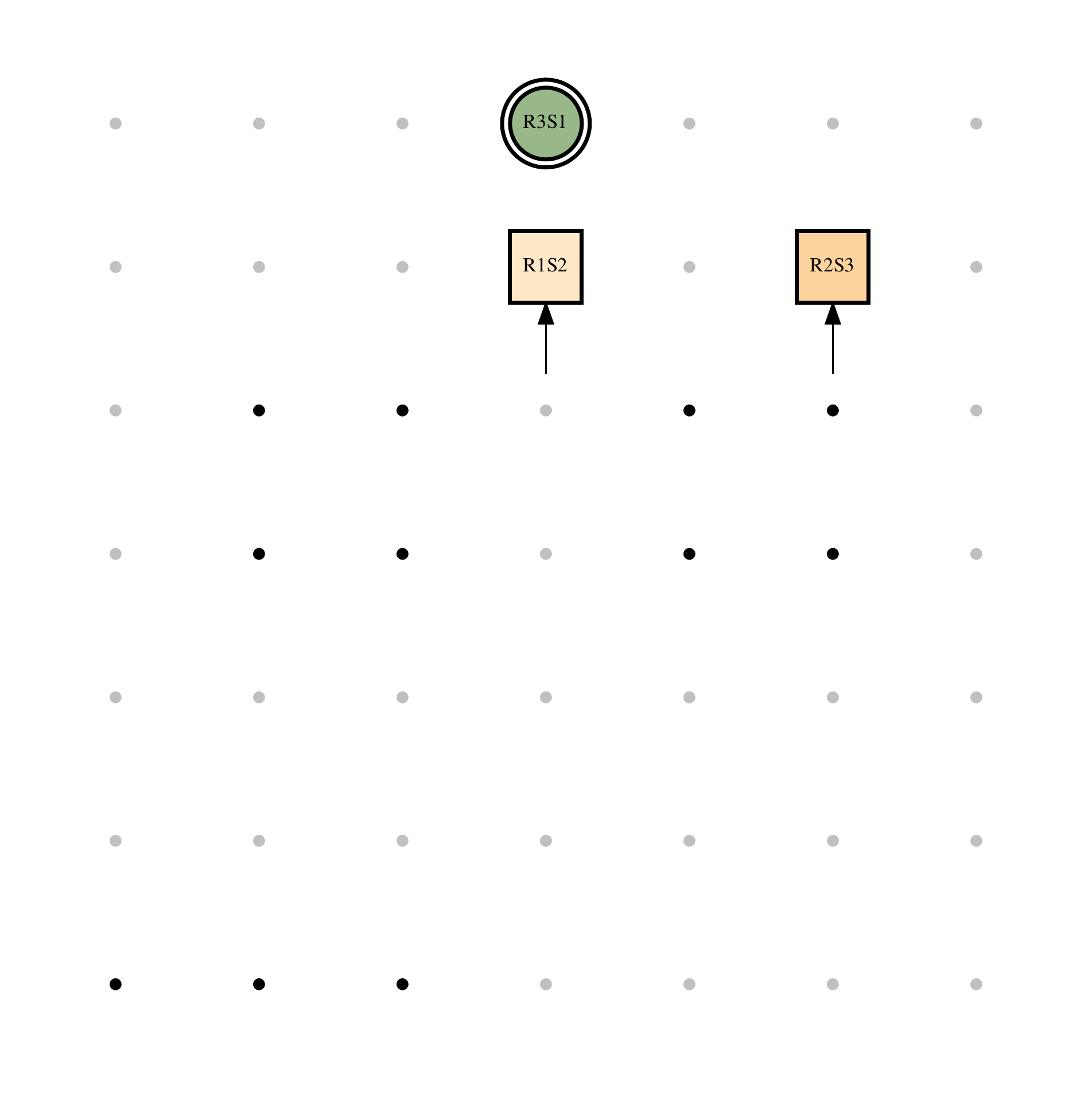}}
  \frame{\includegraphics[scale=0.14]{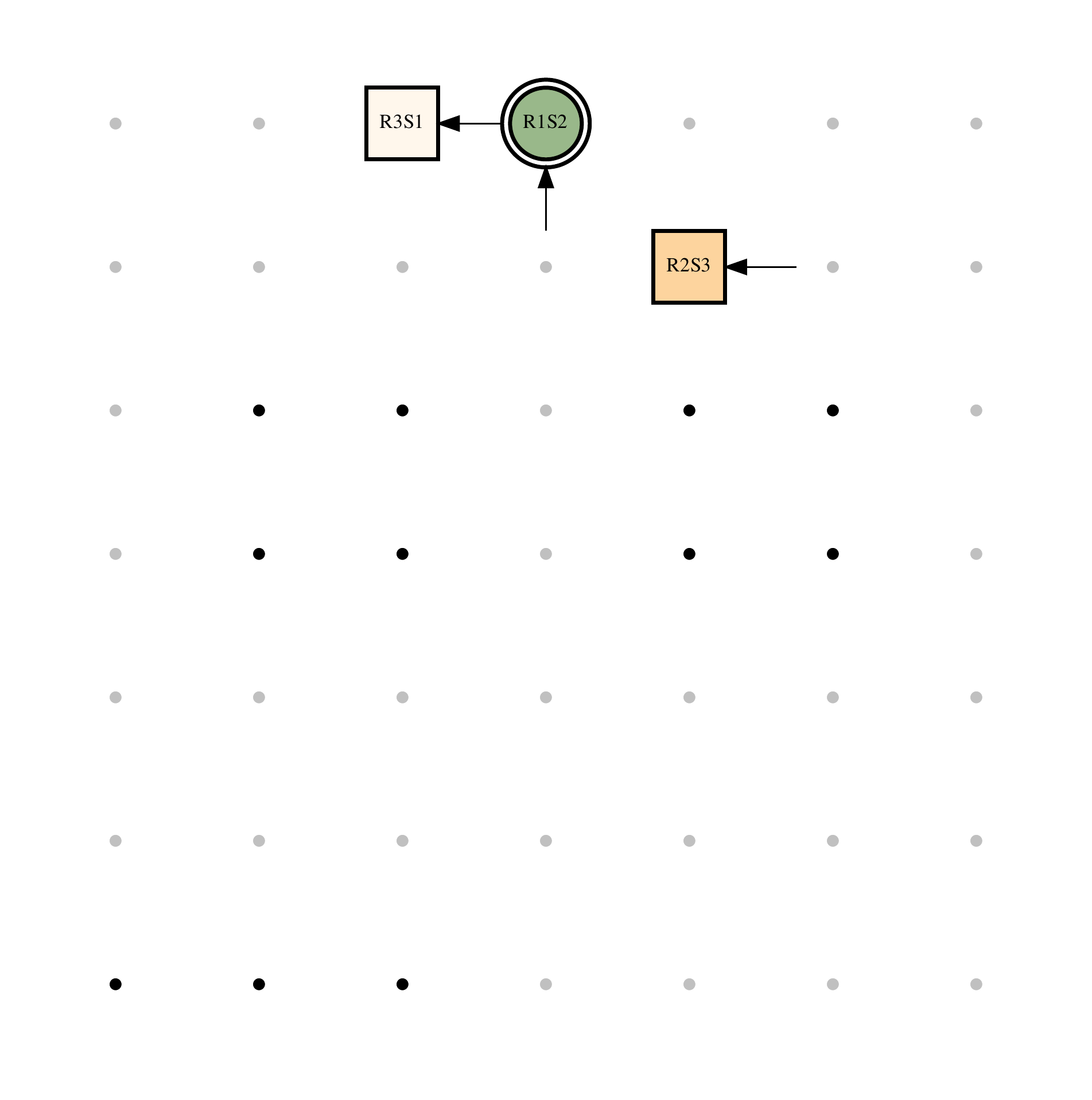}}
  \frame{\includegraphics[scale=0.14]{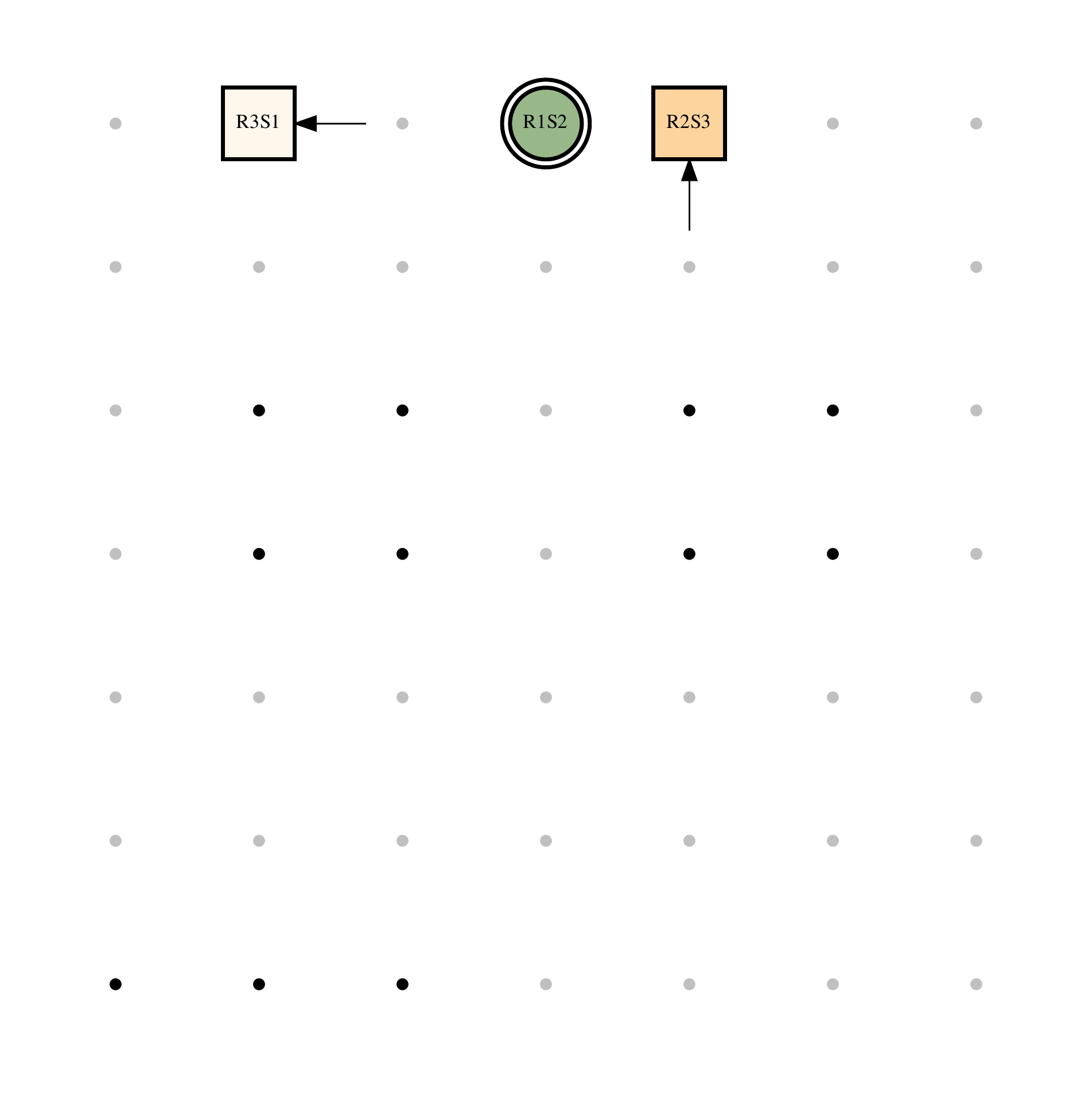}}
  \frame{\includegraphics[scale=0.14]{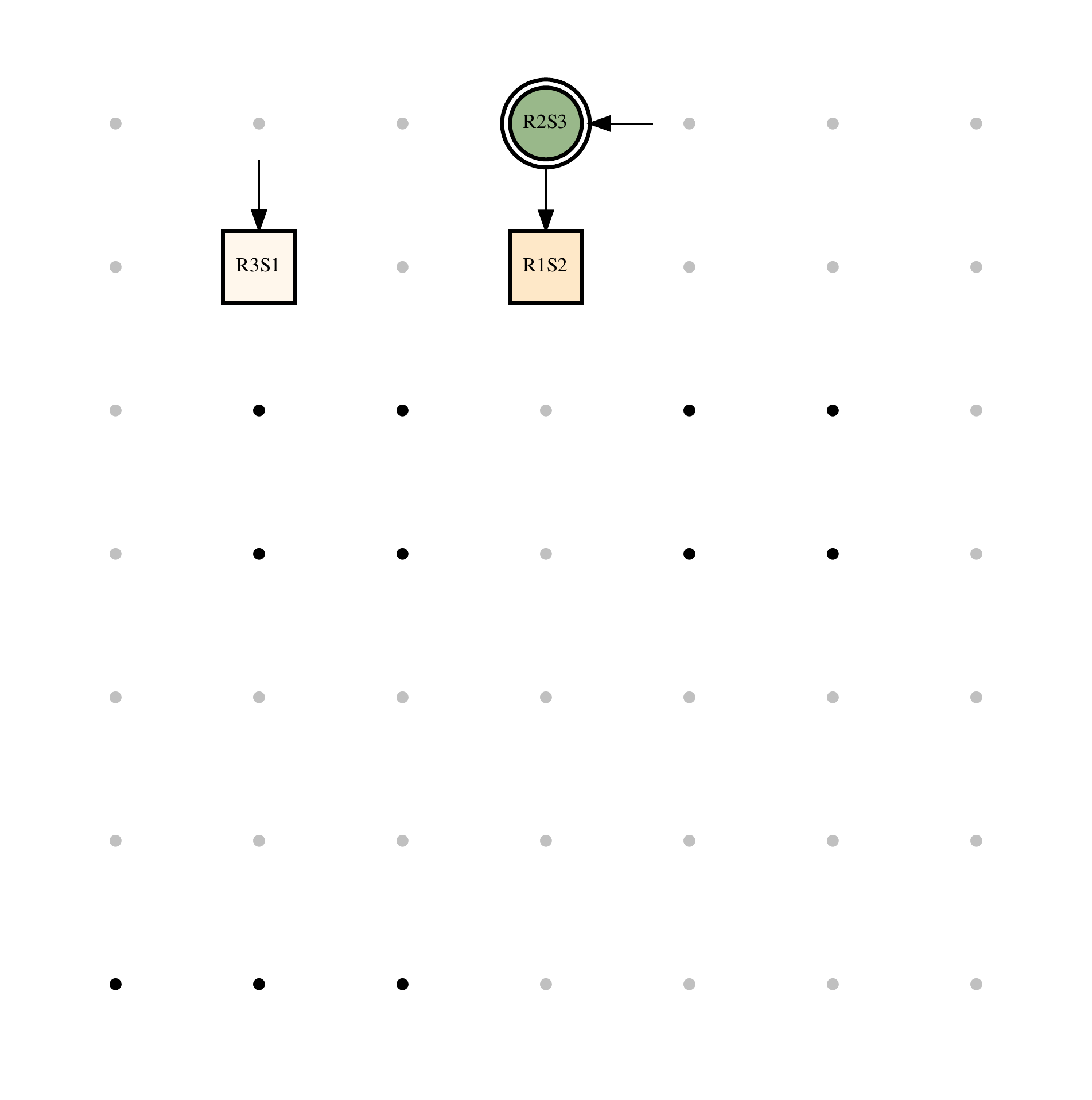}}
  \frame{\includegraphics[scale=0.14]{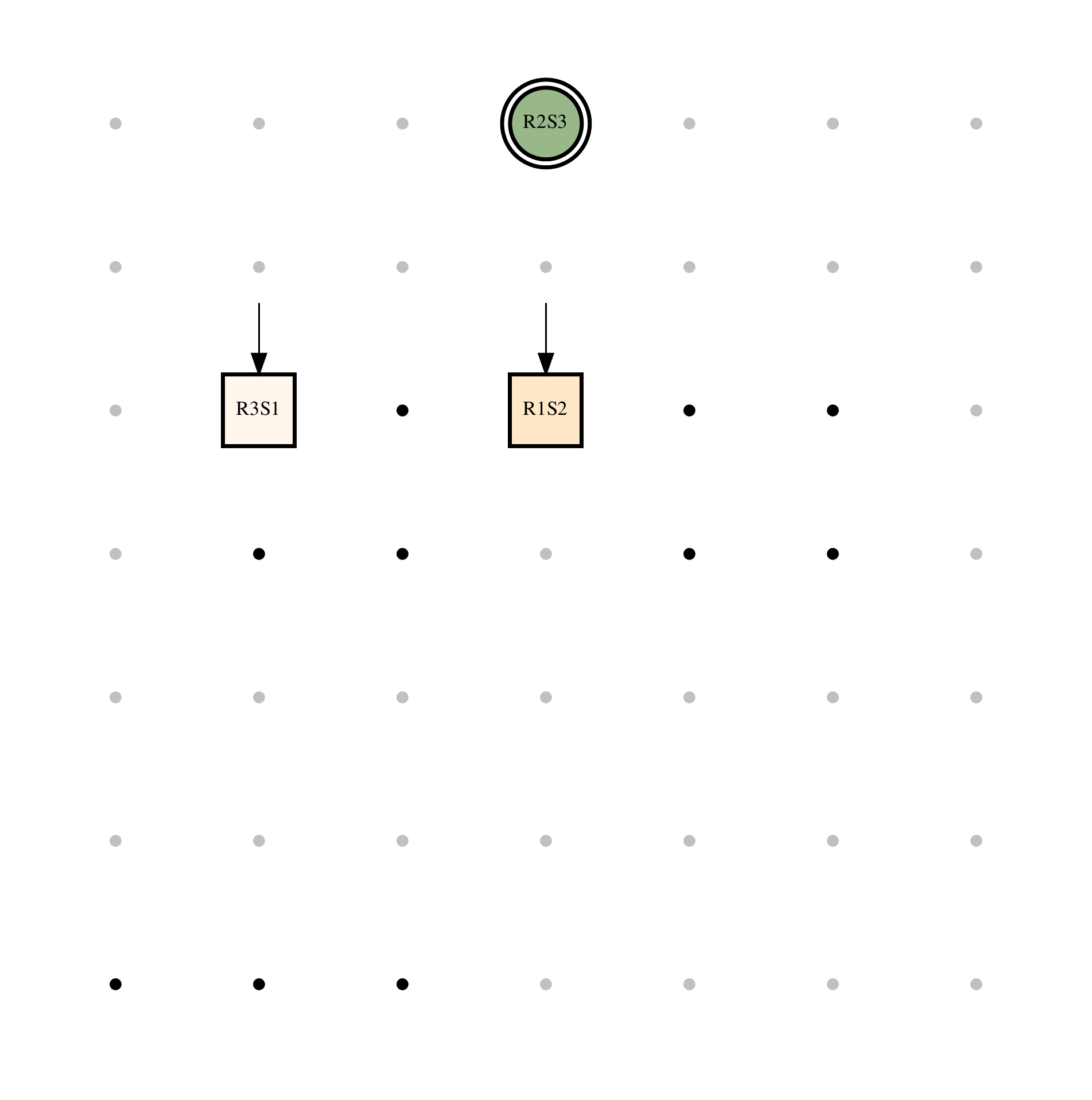}}
  \frame{\includegraphics[scale=0.14]{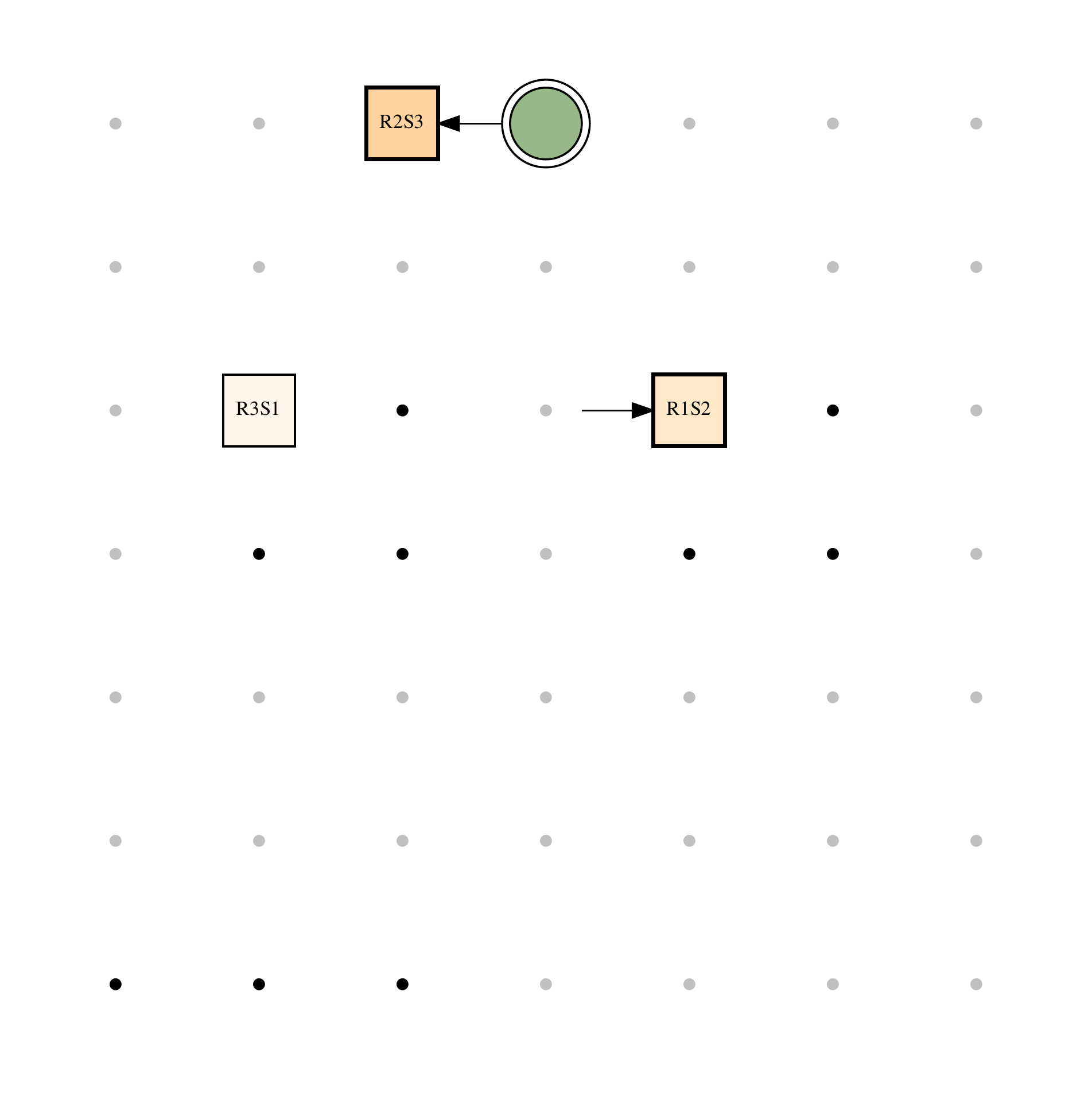}}
  \frame{\includegraphics[scale=0.14]{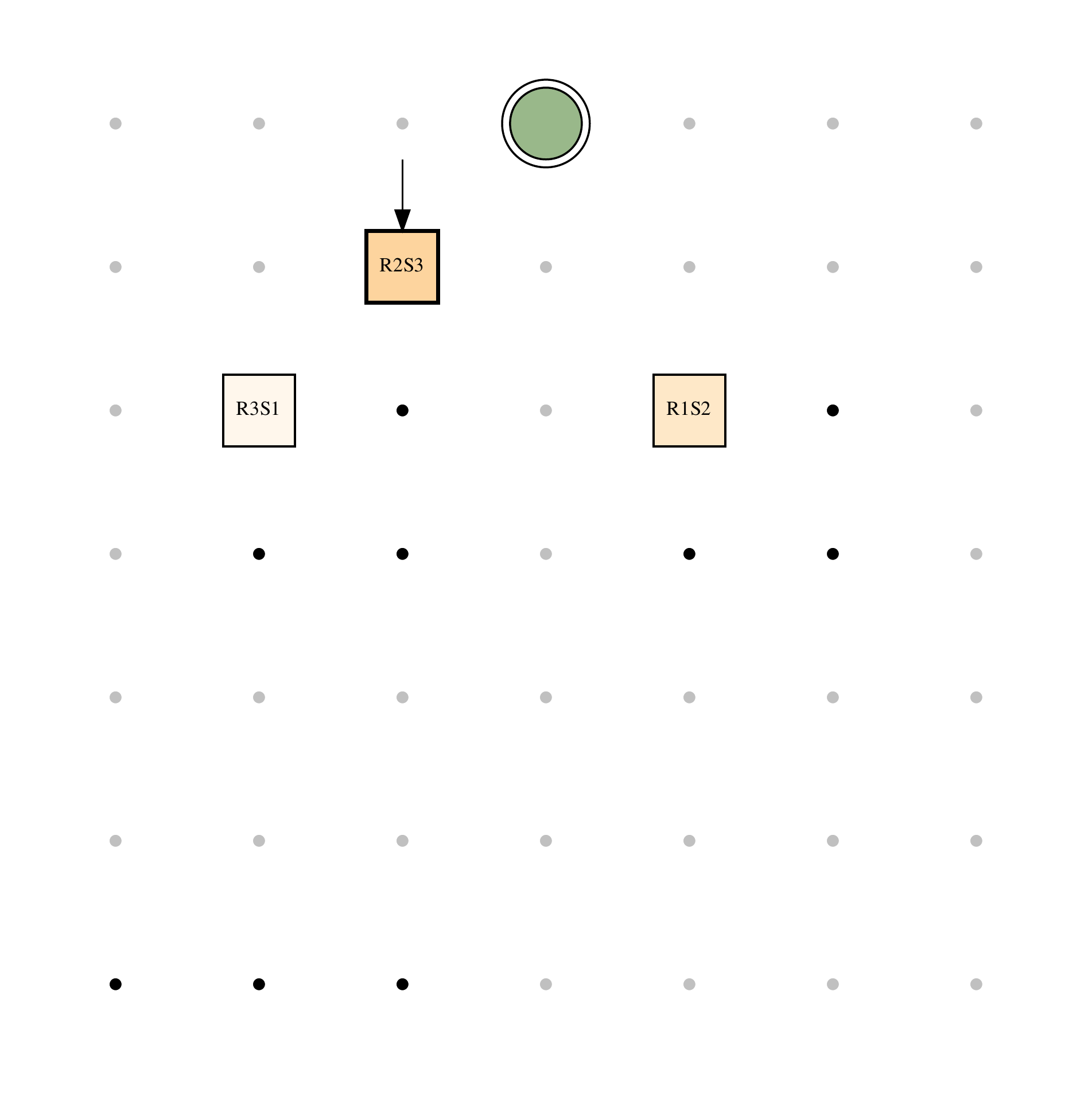}}
  \frame{\includegraphics[scale=0.14]{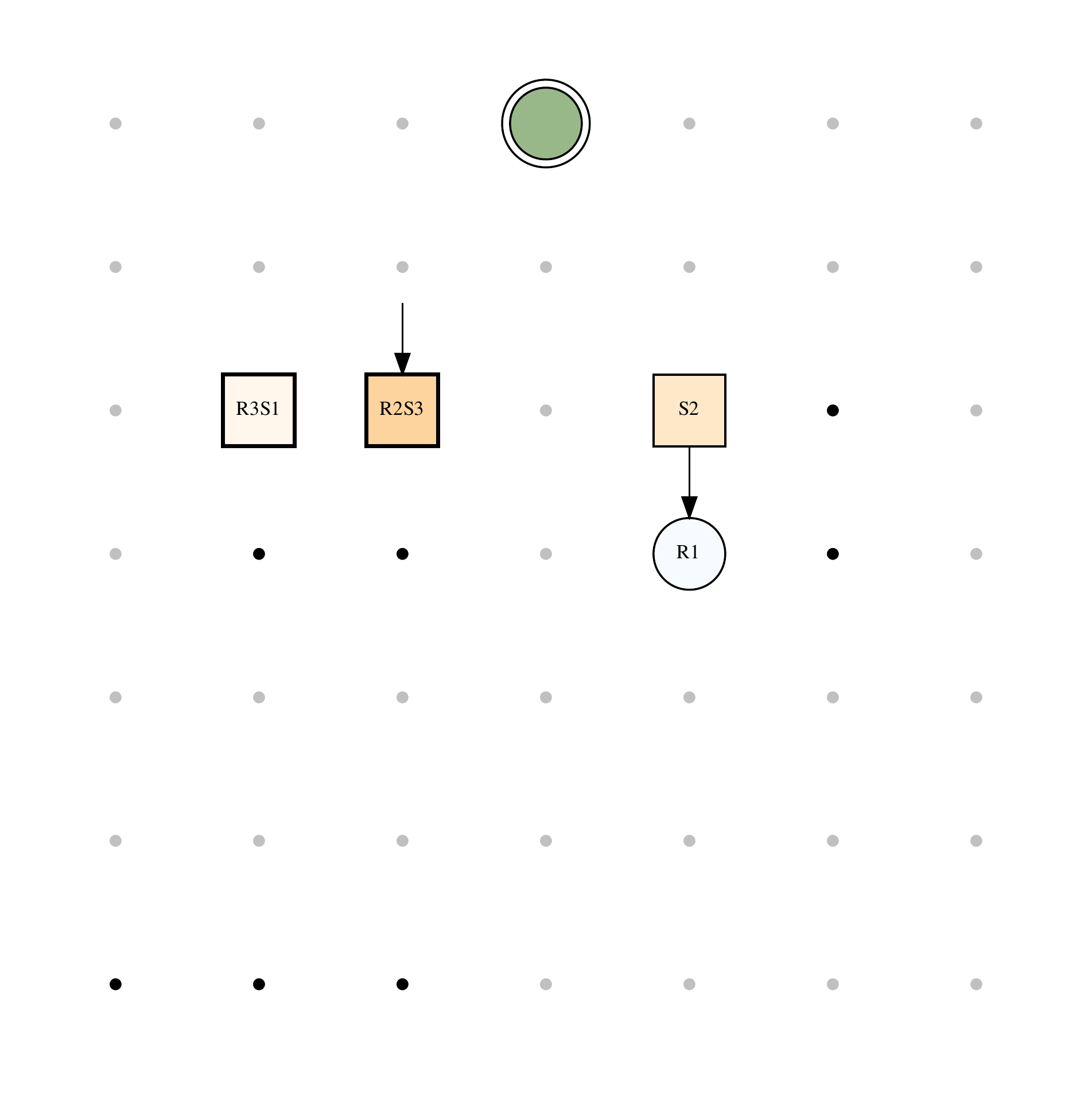}}
  \frame{\includegraphics[scale=0.14]{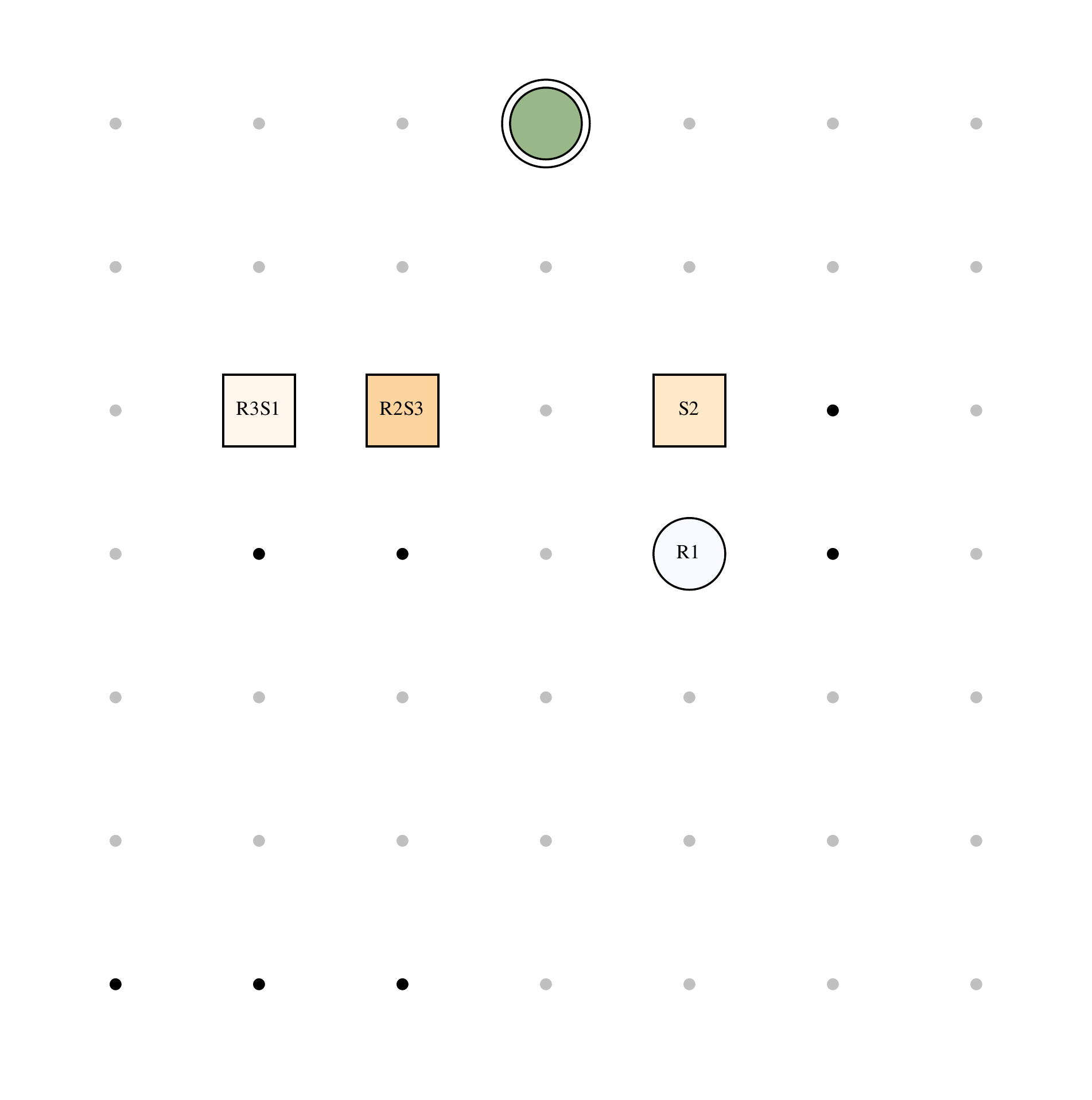}}
  \caption{Individual graph representations making up an animated plan}
  \label{fig:animation}
\end{figure}

Having settled the representation of graphs along with their nodes and edges,
the rest of the visualization encoding mainly deals with setting their attributes.
\begin{table}[ht]
\begin{lstlisting}[language=clingo,basicstyle=\scriptsize\ttfamily,firstnumber=10,linerange={10-10}]
free(P,T) :- not position(_,P,T), position(P), step(T).%%#(\label{lst:asprilo:free}#)
\end{lstlisting}
\begin{lstlisting}[language=clingo,basicstyle=\scriptsize\ttfamily,firstnumber=12,linerange={12-14}]
occo(P,T,robot(R)) :- position(robot(R),P,T),%%#(\label{lst:asprilo:occupied:only:robot:begin}#)
    not position(station(_),P,T),
    not position(shelf(_),P,T).%%#(\label{lst:asprilo:occupied:only:robot:end}#)
\end{lstlisting}
\begin{lstlisting}[language=clingo,basicstyle=\scriptsize\ttfamily,firstnumber=19,linerange={19-19}]
graph(T) :- step(T).%%#(\label{lst:asprilo:graph}#)
\end{lstlisting}
\begin{lstlisting}[language=clingo,basicstyle=\scriptsize\ttfamily,firstnumber=27,linerange={27-27}]
node((P,T),T) :- position(P), step(T).%%#(\label{lst:asprilo:node}#)
\end{lstlisting}
\begin{lstlisting}[language=clingo,basicstyle=\scriptsize\ttfamily,firstnumber=30,linerange={30-31}]
edge((((X,Y),T),((X+DX,Y+DY),T)),T) :- move(robot(R),(DX,DY),T),%%#(\label{lst:asprilo:edge:one}#)
    position(robot(R),(X,Y),T-1).%%#(\label{lst:asprilo:edge:two}#)
\end{lstlisting}
\begin{lstlisting}[language=clingo,basicstyle=\scriptsize\ttfamily,firstnumber=39,linerange={39-44}]
attr(node,(P,T),label,"R{{robot}}") :- position(robot(R),P,T), not position(shelf(_),P,T).%%#(\label{lst:asprilo:label:robot}#)
attr(node,(P,T),label,"S{{shelf}}") :- not position(robot(_),P,T), position(shelf(S),P,T).%%#(\label{lst:asprilo:label:shelf}#)
attr(node,(P,T),label,"R{{robot}}S{{shelf}}") :- position(robot(R),P,T), %%#(\label{lst:asprilo:label:robot-shelfone}#)
    position(shelf(S),P,T).%%#(\label{lst:asprilo:label:robot-shelftwo}#)
attr(node,(P,T),(label,robot),R) :- position(robot(R),P,T).%%#(\label{lst:asprilo:label:robotvar}#)
attr(node,(P,T),(label,shelf),S) :- position(shelf(S),P,T). %%#(\label{lst:asprilo:label:shelfvar}#)
\end{lstlisting}
\begin{lstlisting}[language=clingo,basicstyle=\scriptsize\ttfamily,firstnumber=47,linerange={47-47}]
attr(node,(P,T),shape,"point") :- free(P,T).%%#(\label{lst:asprilo:free:shape}#)
\end{lstlisting}
\begin{lstlisting}[language=clingo,basicstyle=\scriptsize\ttfamily,firstnumber=50,linerange={50-50}]
attr(node,(P,T),shape,"circle") :- occo(P,T,robot(_)).%%#(\label{lst:asprilo:robot:shape}#)
\end{lstlisting}
\begin{lstlisting}[language=clingo,basicstyle=\scriptsize\ttfamily,firstnumber=53,linerange={53-53}]
attr(node,(P,T),color,white) :- free(P,T).%%#(\label{lst:asprilo:free:color}#)
\end{lstlisting}
\begin{lstlisting}[language=clingo,basicstyle=\scriptsize\ttfamily,firstnumber=59,linerange={59-60}]
attr(node,(P,T),colorscheme,"blues9") :- occo(P,T,robot(_)).%%#(\label{lst:asprilo:robot:colorscheme}#)
attr(node,(P,T),fillcolor,R) :- occo(P,T,robot(R)).%%#(\label{lst:asprilo:robot:fillcolor}#)
\end{lstlisting}
  \caption{Selected lines from the visualization encoding for an \asprilo\ scenario (\texttt{viz-asprilo.lp})}
  \label{lst:asprilo:viz}
\end{table}
To see this, consider Table~\ref{lst:asprilo:viz}, giving excerpts of the actual visualization encoding
(using line numbers in the full encoding; lines in between have been dropped for brevity).\footnote{\url{https://github.com/potassco/clingraph/tree/master/examples/asprilo}}
The definition of graphs, nodes, and edges is given in
Line~\ref{lst:asprilo:graph}, Line~\ref{lst:asprilo:node}, and Line~\ref{lst:asprilo:edge:one}-\ref{lst:asprilo:edge:two}.
Let us discuss the remaining lines of interest of \texttt{viz-asprilo.lp} by inspecting some features of a visualization,
produced as follows.
\begin{lstlisting}[basicstyle=\small\ttfamily,numbers=none,xleftmargin=\parindent]
clingo asprilo.lp instance.lp -c horizon=19 --outf=2   | \
clingraph --viz-encoding=viz-asprilo.lp --engine=neato  \
          --out=animate  --sort=asc-int                 \
          --select-model=0 --type=digraph
\end{lstlisting}
The initial call to \clingo\ takes the problem encoding and instance and yields a plan of length 19,
executed on a $7\times 7$ grid with three robots, three shelves, and one picking station.
The individual 20 images underlying the resulting animation are given in Figure~\ref{fig:animation}.
At the beginning,
robots are represented by solid blue circles, shelves by solid orange squares, and the only picking station by a solid
green circle.
This layout changes in the course of the plan.

Let us explain how this works by focusing on unoccupied nodes and robots; shelves and picking stations are treated analogously.
An unoccupied position $p$ at a time step $t$ is captured by \lstinline[mathescape]{free($p$,$t$)} in Line~\ref{lst:asprilo:free}.
Similarly, \lstinline[mathescape]{occo($p$,$t$,$robot(r)$)} tells us that robot $r$ is the only object on position $p$ at a time step $t$.
This is thus neither derivable when a robot is under a shelf, carrying one, or at a picking station.
With this in mind,
we see that Line~\ref{lst:asprilo:free:shape} and~\ref{lst:asprilo:free:color} depict a position as a circle on a white
node (plus omitted details) whenever the position is free.
And analogously,
Line~\ref{lst:asprilo:robot:shape}, \ref{lst:asprilo:robot:colorscheme}, and~\ref{lst:asprilo:robot:fillcolor} represent
solitary robots by solid blue circles.
Here, robots are differentiated using multiple shades of blue via
the \graphviz\ attribute \lstinline{colorscheme},
where each robot selects one color option using an integer in attribute \lstinline{fillcolor}.
Once a robot shares a position with a shelf or picking station, the graphical representation changes
(and instead the robot adopts the one of the shelf or picking station).

Moreover, a robot's label changes whenever it is under a shelf or carries one.
This is handled in Line~\ref{lst:asprilo:label:robot} to~\ref{lst:asprilo:label:shelfvar}.
A template (cf.~Section~\ref{sec:template}) is selected depending on
whether there is a robot, a shelf, or both in the corresponding position (Lines~\ref{lst:asprilo:label:robot}, \ref{lst:asprilo:label:shelf} and \ref{lst:asprilo:label:robot-shelfone} to \ref{lst:asprilo:label:robot-shelftwo}, respectively).
The variables \lstinline{robot} and \lstinline{shelf}, used in the templates, are defined in Line~\ref{lst:asprilo:label:robotvar} and \ref{lst:asprilo:label:shelfvar}, respectively.
 \subsection{Scalable Vector Graphics and interactivity}\label{subsec:svg}

An image format of particular interest is the Scalable Vector Graphics (SVG)\footnote{\url{https://www.w3.org/TR/SVG/Overview.html}}
format as it supports interactivity.
More precisely,
SVG is a text-based web standard for describing images in XML format integrated with Cascading Style Sheets (CSS) and JavaScript.
In order to allow for interactive actions on graphic elements,
we extend the SVG capabilities supported by \graphviz.
Our extension is implemented in JavaScript;
it listens to events being performed on an element and reacts by changing a CSS style property on another element.
To this end,
the \texttt{class} attribute assigned to an element defines how the element changes on a given event:
\lstinline{click}, \lstinline{mouseenter}, \lstinline{mouseleave} and \lstinline{contextmenu} (right click).
Notably, all interaction is single-shot.
That is, the resulting SVG file is generated once by a single call to \clingo\ and
no further interplay with the solver is possible.
Therefore, all information for interactivity needs to be rendered in the same SVG file and
no information of what actions are taken can be returned to the solver.

For illustration, we visualize the mouse-driven expansion of simple trees,
defined by predicates \lstinline{node/1}, \lstinline{parent/2} and \lstinline{root/1}
(see Listing~\ref{lst:mytree} for an example).
The corresponding visualization encoding is given in Listing~\ref{lst:svg-tree}.
\begin{lstlisting}[float,language=clingo,label={lst:svg-tree},basicstyle=\scriptsize\ttfamily,caption={Specification of interaction on trees based on \clingraph's SVG capabilities (\texttt{tree-viz.lp})}]
attr(node,X,style,"filled") :- node(X).%%#(\label{lst:svg-tree:style}#)
edge((X,Y)) :- parent(X,Y).%%#(\label{lst:svg-tree:edge}#)

react(node,X,X) :- node(X).%%#(\label{lst:svg-tree:react1}#)
react(node,Y,X) :- edge((X,Y)).%%#(\label{lst:svg-tree:react2}#)
react(edge,(X,Y),X) :- edge((X,Y)).%%#(\label{lst:svg-tree:react3}#)

attr(node,X,class,@svg_init("visibility","hidden")) :- node(X), not root(X). %%#(\label{lst:svg-tree:initnode}#)
attr(edge,E,class,@svg_init("visibility","hidden")) :- edge(E). %%#(\label{lst:svg-tree:initedge}#)
attr(T,E,class,@svg("click",X,"visibility","visible")) :- react(T,E,X). %%#(\label{lst:svg-tree:click}#)
attr(T,E,class,@svg("mouseenter",X,"opacity","1")) :- react(T,E,X). %%#(\label{lst:svg-tree:enter}#)
attr(T,E,class,@svg("mouseleave",X,"opacity","0.2")) :- react(T,E,X). %%#(\label{lst:svg-tree:leave}#)
\end{lstlisting}
Line~\ref{lst:svg-tree:style} adds a filled style to the nodes and
Line~\ref{lst:svg-tree:edge} generates an edge for each instance of predicate \lstinline{parent/2}.
Lines~\ref{lst:svg-tree:react1} to~\ref{lst:svg-tree:react3} assign the reactions based on the underlying graph element
using atoms of form \lstinline[mathescape]{react($t$,$e_1$,$e_2$)} that are read as:
``element $e_1$ of type $t$ reacts to actions on element $e_2$".
These atoms affect the visibility and opacity of nodes.
For instance,
Line~\ref{lst:svg-tree:react2} tells us that
if there is an edge from  \lstinline{X} to \lstinline{Y} then node \lstinline{Y} reacts to actions on \lstinline{X}.
Lines~\ref{lst:svg-tree:initnode} to~\ref{lst:svg-tree:leave} define the interactivity of the elements by assigning their \lstinline{class} attribute to a formatted string.
This string is handled by our extension,
while the formatting of the strings is done by functions \lstinline{@svg_init(property,value)} and \lstinline{@svg(event,element,property,value)}.
For instance,
in Lines~\ref{lst:svg-tree:initnode} and~\ref{lst:svg-tree:initedge},
function \lstinline{@svg_init} is used to express that node \lstinline{X} and edge \lstinline{E} have
the initial value \lstinline{hidden} for property \lstinline{visibility}.
Line~\ref{lst:svg-tree:click} states that an element~\lstinline{E} changes the value of the CSS property \lstinline{visibility} to \lstinline{visible} when \lstinline{X} is \lstinline{clicked}.
The function \lstinline{@svg} generates the string \lstinline{clicked___X___visibility___visible}
which is assigned as a \lstinline{class} of \lstinline{E} in the SVG file.
The string is then parsed by our extension and mapped into the JavaScript method \lstinline{addEventListener} to react
when \lstinline{X} is clicked.
Similarly, Lines~\ref{lst:svg-tree:enter} and \ref{lst:svg-tree:leave} define that
an element \lstinline{E} changes the value of \lstinline{opacity} to \lstinline{1} or \lstinline{0.2},
whenever the mouse enters or leaves element \lstinline{X},
respectively.

\begin{figure}[ht!]
  \centering
  \frame{\includegraphics[scale=0.17]{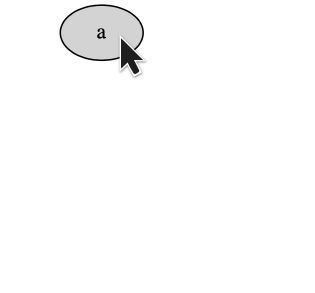}}
  \frame{\includegraphics[scale=0.17]{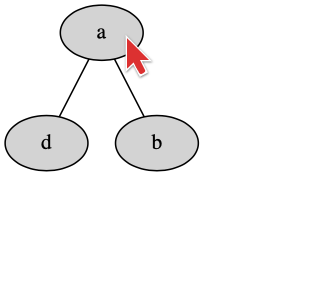}}
  \frame{\includegraphics[scale=0.17]{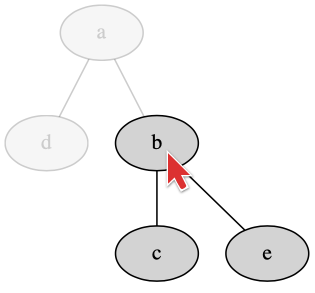}}
  \frame{\includegraphics[scale=0.17]{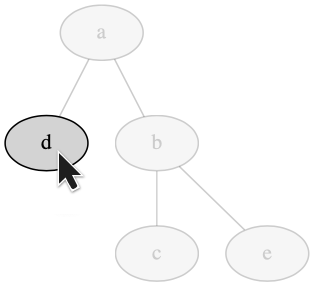}}
  \frame{\includegraphics[scale=0.17]{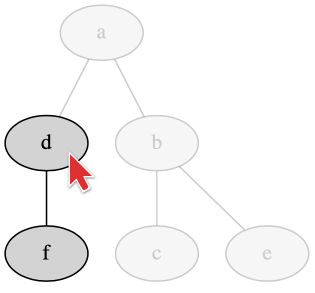}}
  \frame{\includegraphics[scale=0.17]{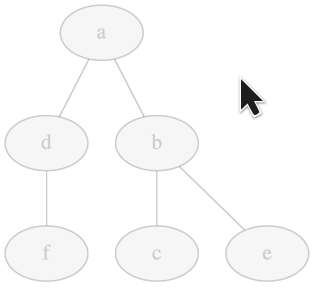}}
  \caption{Example user interaction via mouse actions expanding the tree.}
  \label{fig:svg}
\end{figure}

As an example,
consider the simple tree represented as facts in Listing~\ref{lst:mytree}.
\begin{lstlisting}[float,language=clingos,label={lst:mytree},caption={Facts representing a simple tree (\texttt{mytree.lp})}]
root(a).
node(a). node(b). node(c). node(d). node(e). node(f).
parent(a,b). parent(a,d). parent(b,c). parent(b,e). parent(d,f).
\end{lstlisting}
Together with the visualization encoding in Listing~\ref{lst:svg-tree},
it can be turned into an interactive image in SVG format
by means of the following command:
\begin{lstlisting}[basicstyle=\small\ttfamily,numbers=none,xleftmargin=\parindent]
clingraph mytree.lp --viz-encoding=tree-viz.lp \
          --out=render --format=svg
\end{lstlisting}
A possible user interaction via mouse actions is indicated in Figure~\ref{fig:svg} via a series of snapshots.
While the black pointer highlights positions of interest,
the red one indicates a previous click.
Each such click leads to the expansion of the tree by the succeeding nodes.
Note, how the opacity
of a node and its subnodes changes whenever the pointer hovers over and away from it,
respectively.

The transformation of \graphviz\ into SVG uses the group element \lstinline{<g>} to group all elements related to a node or edge.
Since only these group elements can be indexed in JavaScript,
the CSS style properties are set on the SVG group.
This results in the limitation that
CSS style properties are not overwritten on the elements contained in the group.
Thus, many property changes have no impact.
In particular, this issue leads to problems when changing the color of elements.
To address this, we offer a workaround for changing colors dynamically.
That is,
we provide the function \lstinline{@svg_color} to represent the CSS value \lstinline{currentcolor}.
This can be used with any \graphviz\ color attribute, such as \lstinline{color}, \lstinline{fillcolor}, and
\lstinline{fontcolor},
and serves as a placeholder for the color set using the SVG class.

\begin{figure}[ht!]
  \centering
  \frame{\includegraphics[scale=0.18]{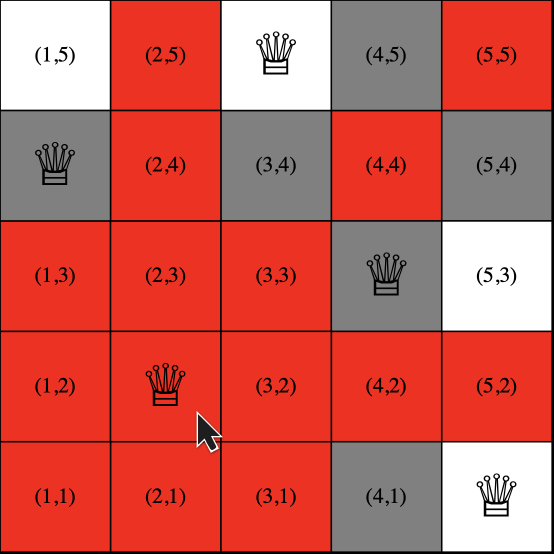}} \quad
  \frame{\includegraphics[scale=0.18]{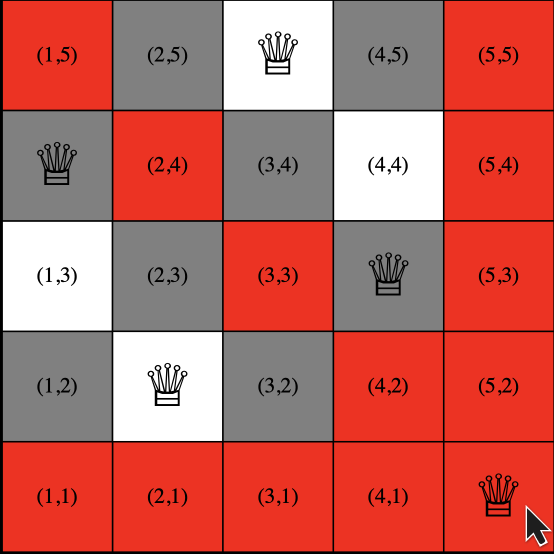}} \quad
  \frame{\includegraphics[scale=0.18]{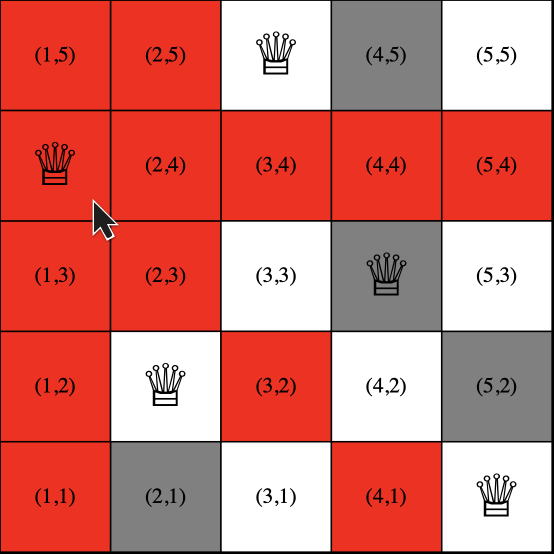}}
  \caption{Three snapshots of the user hovering the mouse pointer over specific cells on the board.}
  \label{fig:svg-queens}
\end{figure}

For illustrating this functionality,
Listing~\ref{lst:svg-queens} extends the queens example with interactive elements
for visualizing all cells attacked by a queen once it is hovered over by the mouse pointer.
\begin{lstlisting}[float,language=clingo,label={lst:svg-queens},basicstyle=\scriptsize\ttfamily,caption={SVG interactive queens (\texttt{svg-queens.lp})}]
attr(node,(X,Y),fillcolor,@svg_color()) :- cell(X,Y).%%#(\label{lst:svg-queens:fillcolor}#)

attr(node,(X,Y),class,@svg_init("color","gray")) :- cell(X,Y), (X+Y)\2 != 0. %%#(\label{lst:svg-queens:initgray}#)
attr(node,(X,Y),class,@svg_init("color","white")) :- cell(X,Y), (X+Y)\2 == 0. %%#(\label{lst:svg-queens:initwhite}#)
attr(node,C,class,@svg("mouseenter",Q,"color","red")) :- attack(Q,C). %%#(\label{lst:svg-queens:enter}#)
attr(node,(X,Y),class,@svg("mouseleave",Q,"color","gray")) :- attack(Q,(X,Y)), (X+Y)\2 != 0. %%#(\label{lst:svg-queens:leavegray}#)
attr(node,(X,Y),class,@svg("mouseleave",Q,"color","white")) :- attack(Q,(X,Y)), (X+Y)\2 == 0. %%#(\label{lst:svg-queens:leavewhite}#)
\end{lstlisting}
To this end,
we replace Lines~\ref{lst:queens:fillcolor:gray} and~\ref{lst:queens:fillcolor:white}
in the visualization encoding in Listing~\ref{lst:queens} by
Line~\ref{lst:svg-queens:fillcolor} in Listing~\ref{lst:svg-queens} to use the color
set by the interactions specified in the following lines.
Lines~\ref{lst:svg-queens:initgray} and~\ref{lst:svg-queens:initwhite} set the initial color of the nodes to
\lstinline{gray} and \lstinline{white}, respectively.
Line~\ref{lst:svg-queens:enter} adds functionality to a node \lstinline{C} attacked by queen \lstinline{Q} so that the
color is set to \lstinline{red} when the mouse enters \lstinline{Q}.
When the mouse leaves \lstinline{Q},
Lines~\ref{lst:svg-queens:leavegray} and \ref{lst:svg-queens:leavewhite} set the color back to the original value.
We illustrate this by means of three snapshots in Figure~\ref{fig:svg-queens}.
In the one on the left, the mouse pointer hovers over position (2,2).
Accordingly, all cells on the same row, column, and diagonals are colored in red.
The same happens in the two other scenarios,
though initiated from position (5,1) and (1,4), respectively.

The approach has further limitations.
For example,
labels are independent of a CSS style and thus cannot be changed interactively.
A way around this is to create multiple layers of nodes with the same position and change their visibility.\footnote{A nice example for this is the \emph{minesweeper} puzzle given at \url{https://github.com/potassco/clingraph/tree/master/examples/minesweeper}.}
However, we have no control over which elements appear on top and which on the bottom.
Rather this must be handled manually by assuring that only a single element is visible in each position at each time.
Another issue is that the position of all elements is fixed;
therefore, expanding the size of the image on demand is impossible, only its visibility can be changed.

 \subsection{Visualizing the solving process of a Sudoku puzzle}\label{subsec:vizsolving}

Up to now, all case studies take answer sets as input for visualization.
For the next example, however,
we visualize partial assignments appearing during the search process of \clingo.
Specifically, we discuss a visualization of the solving process of a Sudoku puzzle.
To this end,
we rely on \clingo's capacity of integrating user-defined propagators\footnote{\url{https://potassco.org/clingo/python-api/current/clingo/propagator.html}}
into the
solving process and use \clingraph's API for
streamlining the declarative visualization of partial assignments.

\begin{table}[p]
  \begin{minipage}{1.0\linewidth}
\begin{lstlisting}[language=python,escapeinside={\#\#\%}{\#},basicstyle=\scriptsize\ttfamily,firstnumber=8,linerange={9-77}]
class ClingraphPropagator:
    def __init__(self, viz_encoding):
        self.viz_encoding = viz_encoding##%\label{lst:prop:init:start}#
        self.factbases = []
        self.l2s = {}##%\label{lst:prop:init:end}#

    def init(self, init):##%$\;\;\;...$\label{lst:prop:init}#
        for atom in init.symbolic_atoms:
            lit = init.solver_literal(atom.literal)
            self.l2s.setdefault(lit, []).append(str(atom.symbol))##%\label{lst:prop:init:l2smap}#
            init.add_watch(lit)##%\label{lst:prop:init:watch}#

    def propagate(self, ctl, changes):##%\label{lst:prop:propagate}#
        i = len(self.factbases)
        propagation_prg = [f"_step_type(propagate,{i}).", f"_level({ctl.assignment.decision_level})."]##%\label{lst:prop:steplevel}#
        for l,symbols in self.l2s.items():##%\label{lst:prop:parassign:start}#
            v = ctl.assignment.value(l)##%\label{lst:prop:parassign:query}#
            t = '_undefined' if v is None else '_true' if v else '_false'
            for s in symbols:
                propagation_prg.append(f"{t}({s}).")##%\label{lst:prop:parassign:end}#
        for l in changes:##%\label{lst:prop:changes:start}#
            symbols = self.l2s[l]
            for s in symbols:
                propagation_prg.append(f"_change({s}).")##%\label{lst:prop:changes:end}#
        self.add_factbase(propagation_prg)##%\label{lst:prop:addfb:call}#
        return True##%\label{lst:prop:propagate:end}#

    def undo(self, solver_id, assign, undo):##%\label{lst:prop:undo}#
        i = len(self.factbases)
        propagation_prg = [f"_step_type(undo,{i}).", f"_level({assign.decision_level})."]##%\label{lst:prop:undo:steplevel}#

        for l,symbols in self.l2s.items():##%\label{lst:prop:undo:parassign:start}#
            v = assign.value(l)
            t = '_undefined' if v is None else '_true' if v else '_false'
            for s in symbols:
                propagation_prg.append(f"{t}({s}).")##%\label{lst:prop:undo:parassign:end}#
        for l in undo:##%\label{lst:prop:undo:changes:start}#
            symbols = self.l2s[l]
            for s in symbols:
                propagation_prg.append(f"_change({s}).")##%\label{lst:prop:undo:changes:end}#

        self.add_factbase(propagation_prg)

    def decide(self, thread_id, assign, fallback):##%\label{lst:prop:decide}#
        i = len(self.factbases)
        propagation_prg = [f"_step_type(decide,{i}).", f"_level({assign.decision_level})."]##%\label{lst:prop:decide:steplevel}#

        for l,symbols in self.l2s.items():##%\label{lst:prop:decide:parassign:start}#
            v = assign.value(l)
            t = '_undefined' if v is None else '_true' if v else '_false'
            for s in symbols:
                propagation_prg.append(f"{t}({s}).")##%\label{lst:prop:decide:parassign:end}#
        if abs(fallback) in self.l2s:##%\label{lst:prop:decide:start}#
            for s in self.l2s[abs(fallback)]:
                pol = "pos" if fallback > 0 else "neg"
                propagation_prg.append(f"_decide({s},{pol}).")##%\label{lst:prop:decide:end}#

        self.add_factbase(propagation_prg)
        return 0

    def add_factbase(self, prg_list):##%\label{lst:prop:addfb}#
        fb = Factbase()
        ctl = Control([])
        ctl.load(self.viz_encoding)
        ctl.add("base",[],"".join(prg_list))
        ctl.ground([("base",[])],ClingraphContext())
        ctl.solve(on_model=fb.add_model)
        self.factbases.append(fb)##%\label{lst:prop:addfb:end}#
\end{lstlisting}
\end{minipage}
\caption{The propagator class for visualizing solving}
  \label{lst:prop}
\end{table}

In Table~\ref{lst:prop}, we provide a generic propagator
that can be used
directly to monitor solving or
as a template to create a domain-specific propagator.
Basically, \lstinline{ClingraphPropagator} class implements the
user-defined propagator interface expected by \clingo's Python API.
Its instance variables are initialized in
Lines~\ref{lst:prop:init:start} to \ref{lst:prop:init:end}.
The \lstinline{viz_encoding} variable holds the path of
the visualization encoding specific to the problem domain.
The propagator uses this encoding to generate the
facts defining the graph to visualize each
partial assignment, which are stored in the \lstinline{factbases} list.
Additionally, the instance variable \lstinline{l2s}
maps each literal used internally by \clingo\ to
the corresponding list of atomic symbols from the problem encoding.
Specifically, this mapping is formed in Line~\ref{lst:prop:init:l2smap},
just before solving process starts, when \clingo\ calls the \lstinline{init}
function (Line~\ref{lst:prop:init}) of \lstinline{ClingraphPropagator}.
Note that in Line~\ref{lst:prop:init:watch} the propagator requests
the solver to be notified when the truth value of these internal
literals changes.
Hence, with the help of \lstinline{l2s},
the propagator functions can find the corresponding atoms of
a solver literal whose truth value has changed during solving.

The main functionality of the propagator is to compile and prepare
partial assignments appearing during various stages of the search process
as reified atoms,
which are passed to the visualization encoding.
Such facts are
of the form
\lstinline[mathescape]{_true($a$)},
\lstinline[mathescape]{_false($a$)} and
\lstinline[mathescape]{_undefined($a$)} for each atom $a$ if it is
assigned to true, false or neither
in the current partial assignment,
respectively.
The key stages account for times when \clingo\
reaches a fixpoint during unit propagation;
decides on a literal; or
faces a conflict and is about to backtrack.
In each situation,
\clingo\ calls the corresponding propagator function
\lstinline{propagate}~(Line~\ref{lst:prop:propagate}),
\lstinline{decide} (Line~\ref{lst:prop:decide}) or
\lstinline{undo} (Line~\ref{lst:prop:undo}), respectively,
and makes the partial assignment accessible to them.
Hence, these functions are suitable for preparing the reified atoms
of the partial assignment at the time of the call.
In the \lstinline{propagate} function, for instance,
these facts are generated in
Lines~\ref{lst:prop:parassign:start} to \ref{lst:prop:parassign:end} and
functions \lstinline{decide} and \lstinline{undo} have the same corresponding
statements.
Note that for each solver literal, corresponding atoms are found via
the mapping \lstinline{l2s} and
truth values of such atoms are queried in the current partial assignment in
Line~\ref{lst:prop:parassign:query}.
Additionally,
in each stage
we generate the fact \lstinline[mathescape]{_step_type($t$,$i$)}
where $t$ is either \lstinline{propagate}, \lstinline{decide} or \lstinline{undo},
and
$i$ is a natural number identifying the solving step
(in Lines~\ref{lst:prop:steplevel}, \ref{lst:prop:undo:steplevel} and \ref{lst:prop:decide:steplevel}).
Such facts are required not only to designate the type of the current stage,
but also to order the visualization of each generated partial assignment.
This ordering allows us to represent \clingo's solving process
by combining individual graphs as an animation.
The functions \lstinline{propagate} and \lstinline{undo}
generate additional facts of the form \lstinline[mathescape]{_change($a$)},
where the truth value assignment to atom $a$ has changed during the propagation (Lines~\ref{lst:prop:changes:start} to \ref{lst:prop:changes:end})
or will be undone during backtracking (Lines~\ref{lst:prop:undo:changes:start} to \ref{lst:prop:undo:changes:end}), respectively.
Similarly,
the function \lstinline{decide} generates a reified fact of the form
\lstinline[mathescape]{_decide($a$,$p$)} in Lines~\ref{lst:prop:decide:start} to \ref{lst:prop:decide:end}
to represent \clingo\ chooses atom $a$ to have a truth value of true or false
depending on $p$ being \lstinline{pos} or \lstinline{neg},
at a decision point, respectively.

For each solving stage,
we process the reified atoms of the active partial assignment
with the problem domain's visualization encoding
to generate the facts defining the graph.
This is achieved by calling the \lstinline{add_factbase} function
defined in Lines~\ref{lst:prop:addfb} to \ref{lst:prop:addfb:end} at the end of each solving stage.
Each resulting graph facts gets stored in a \texttt{Factbase}
object of \clingraph's API in Line~\ref{lst:prop:addfb:end}.
Once \clingo's solving is done,
we process all \texttt{Factbase} objects accumulated in the propagator
using \clingraph\
to generate individual graphs for each of the partial assignments.
Finally, we combine these graphs to generate an animation of
\clingo's solving process.
Unlike the previous examples,
we rely on \clingraph's API functions
(eg., \lstinline{compute_graphs} and \lstinline{save_gif})
to carry out these tasks.

\begin{figure}[ht!]
  \centering
  \frame{\includegraphics[scale=0.17]{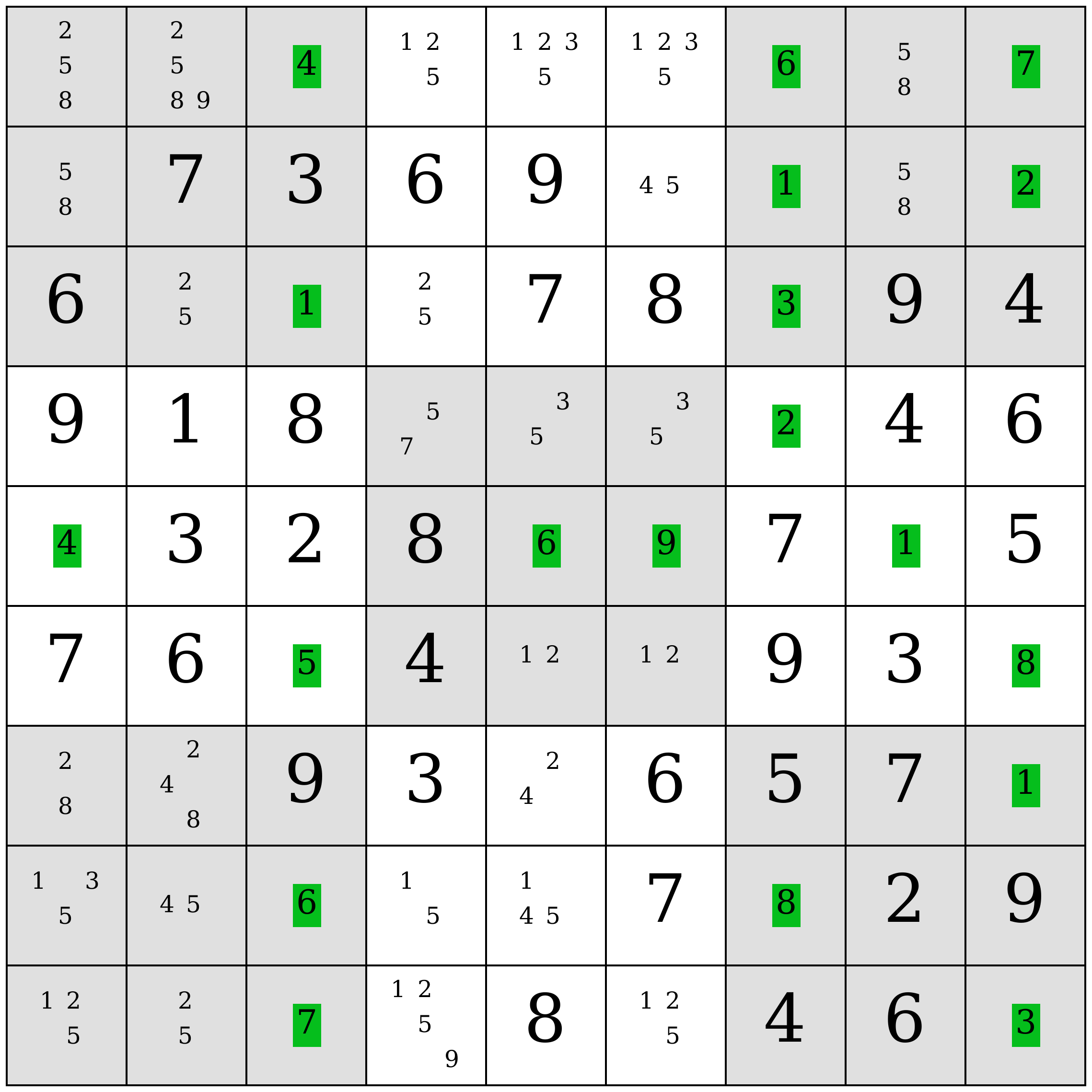}}
  \frame{\includegraphics[scale=0.17]{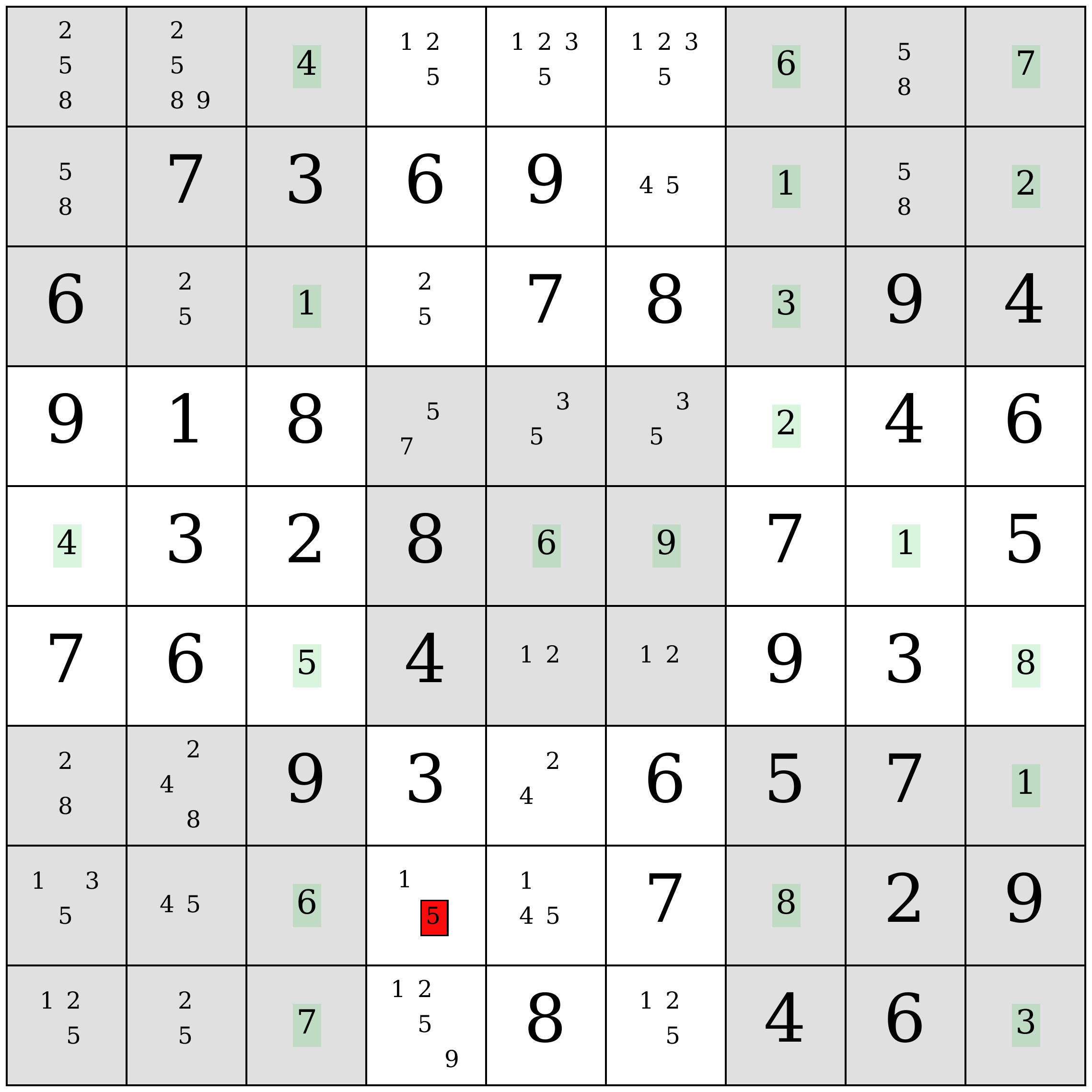}}
  \frame{\includegraphics[scale=0.17]{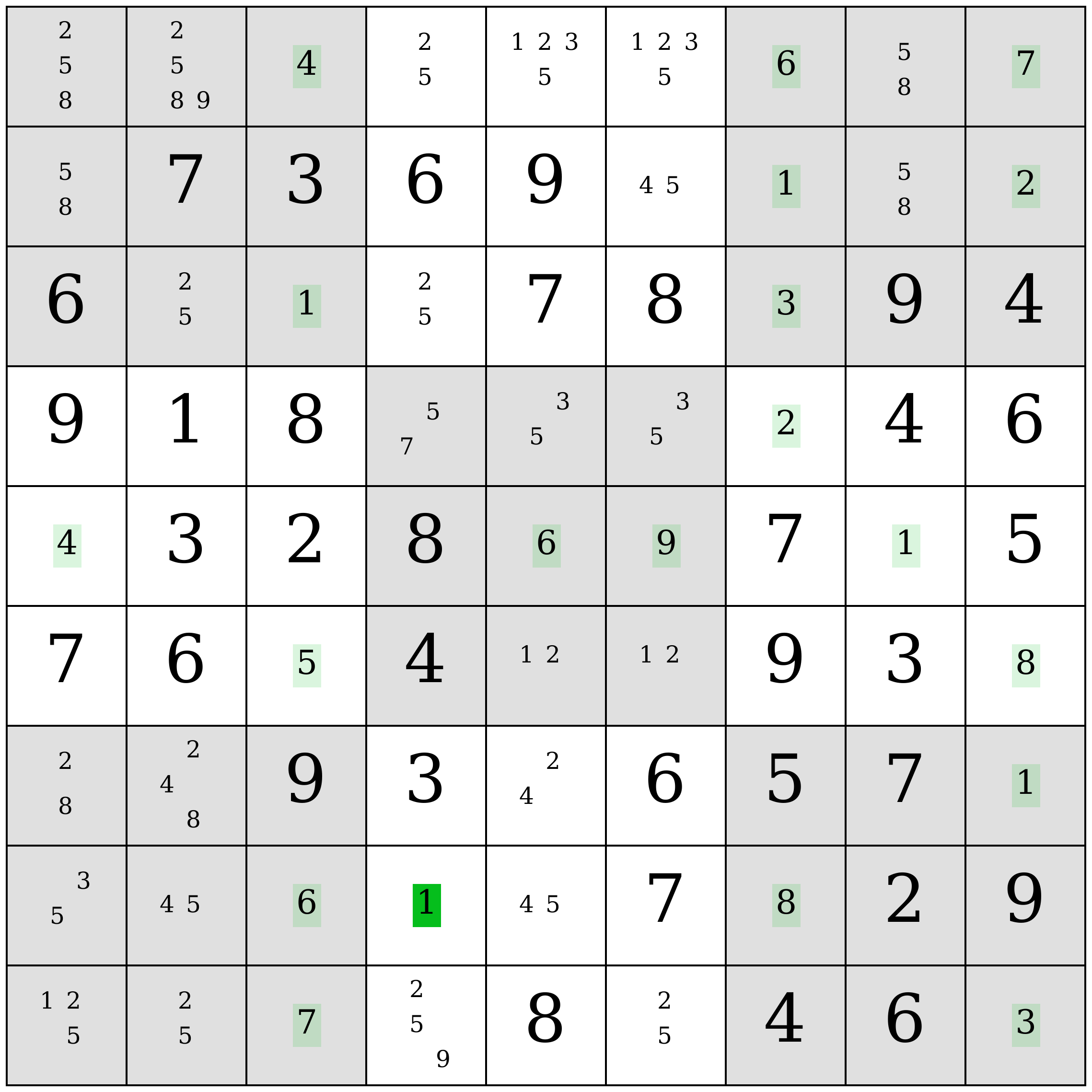}}
  \frame{\includegraphics[scale=0.17]{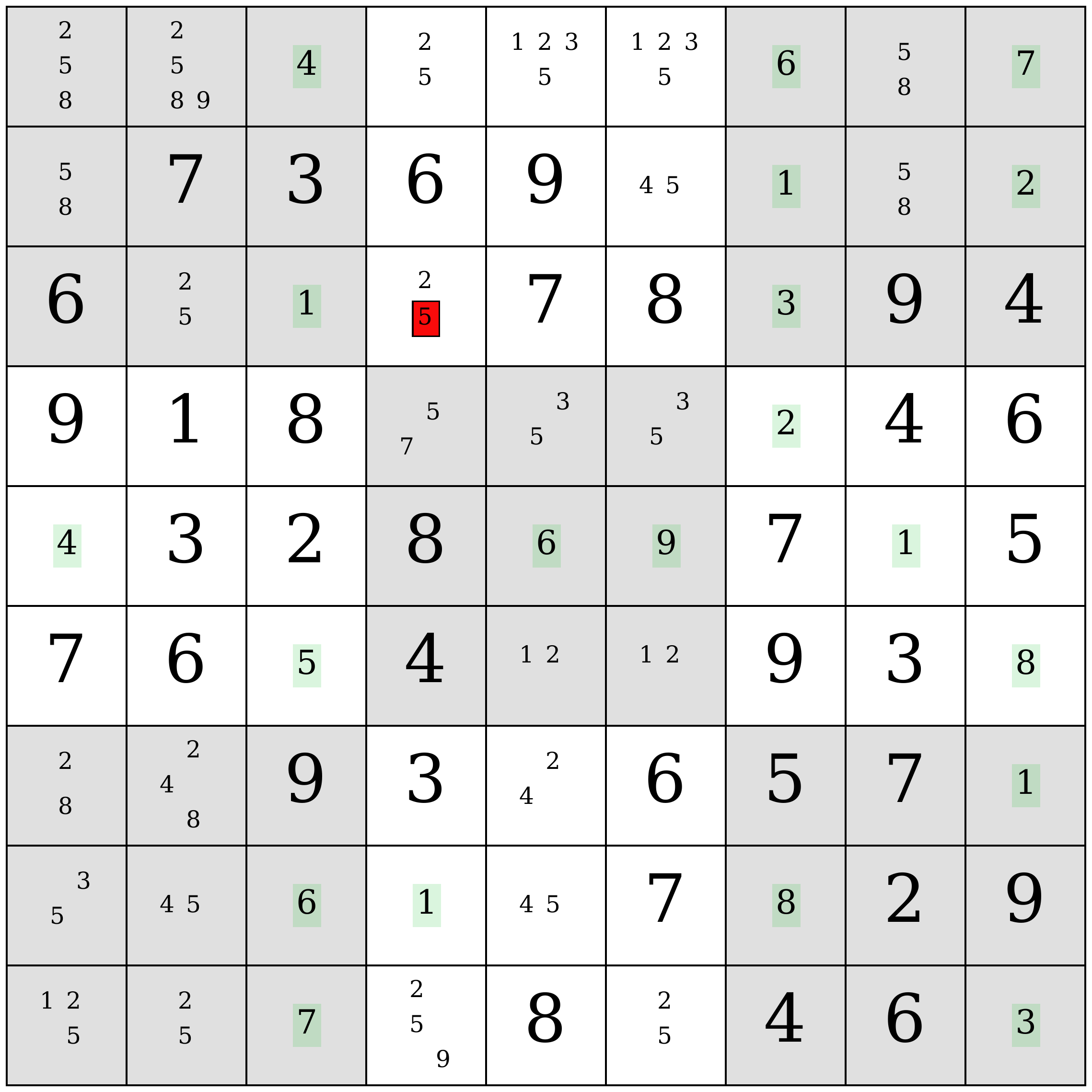}}
  \frame{\includegraphics[scale=0.17]{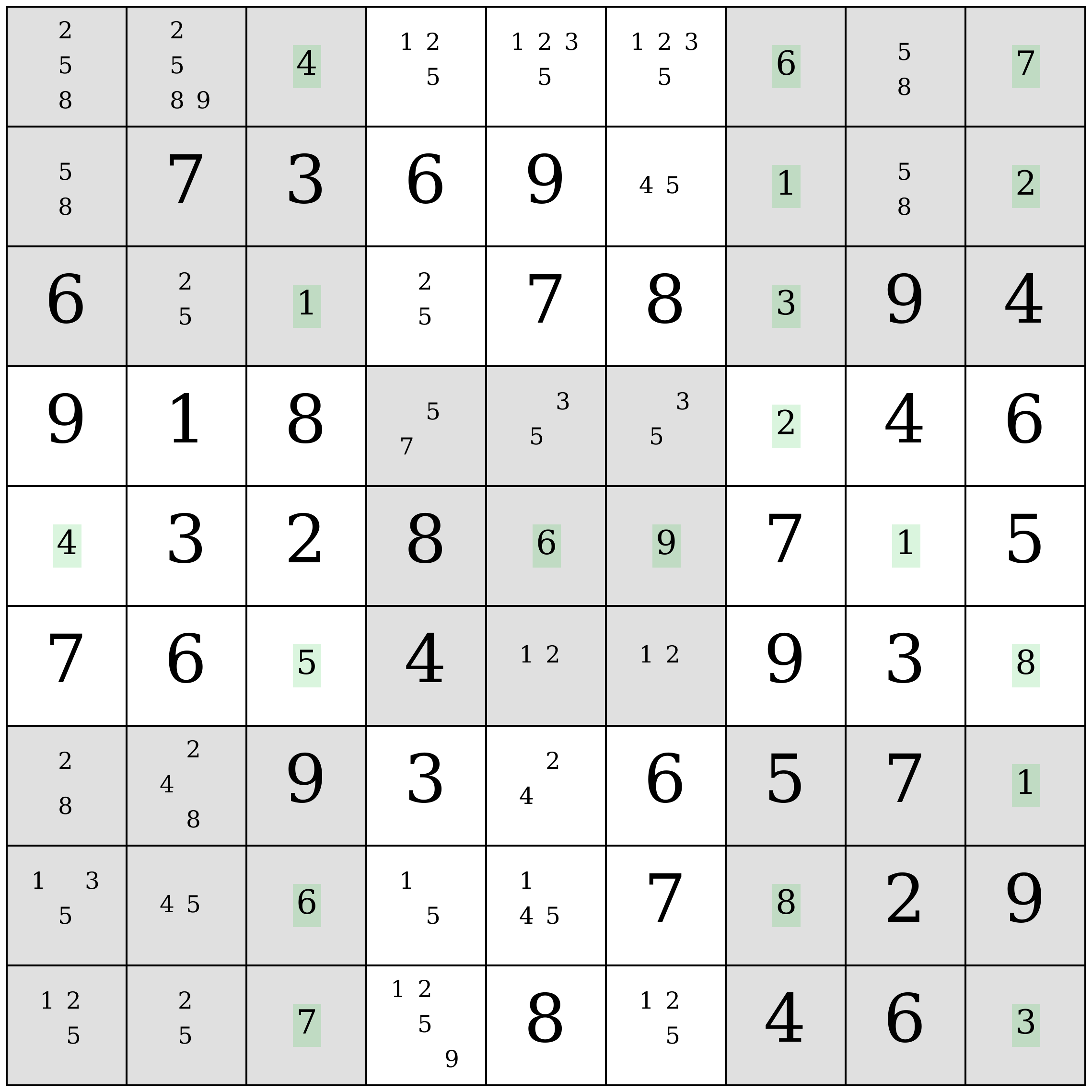}}
  \frame{\includegraphics[scale=0.17]{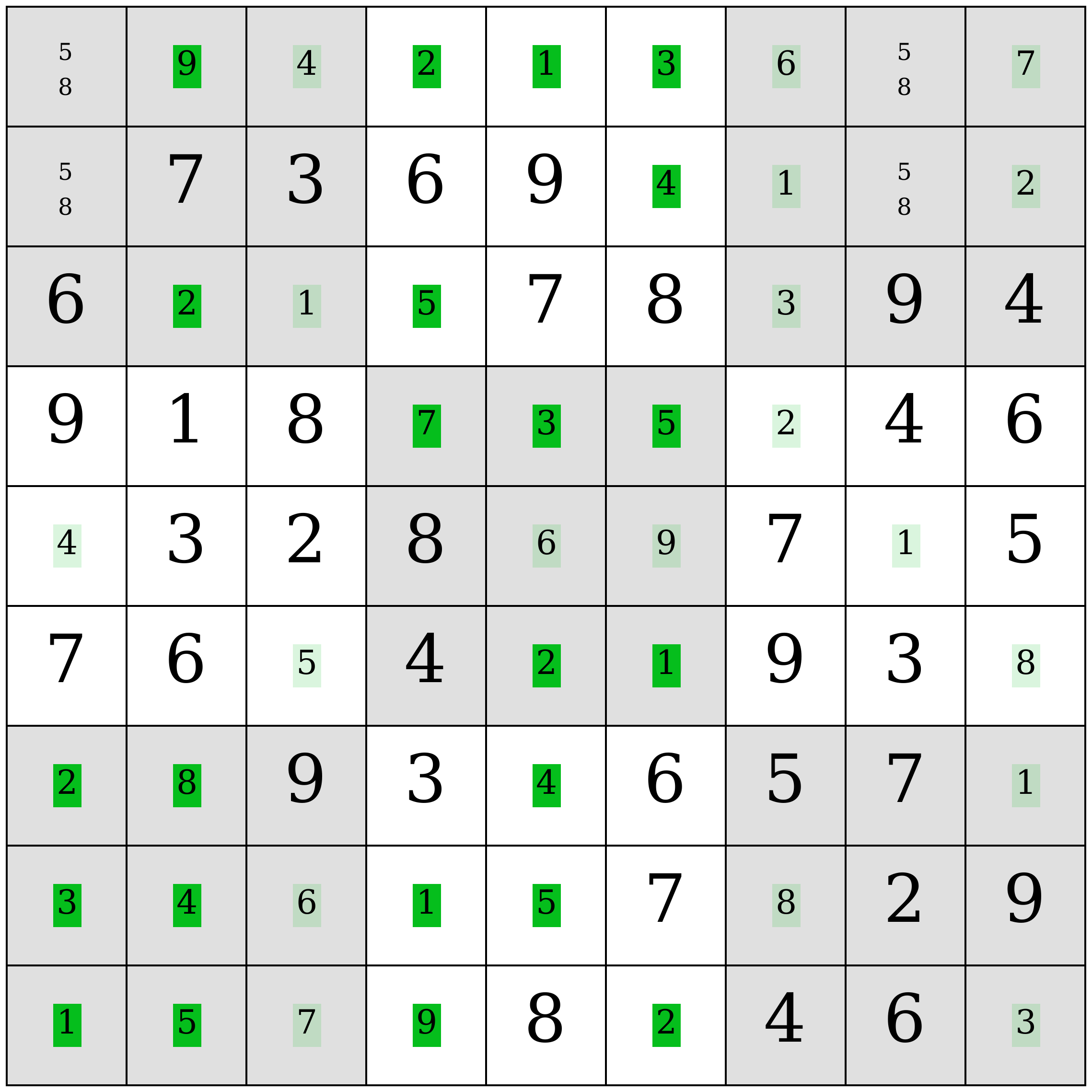}}
  \frame{\includegraphics[scale=0.17]{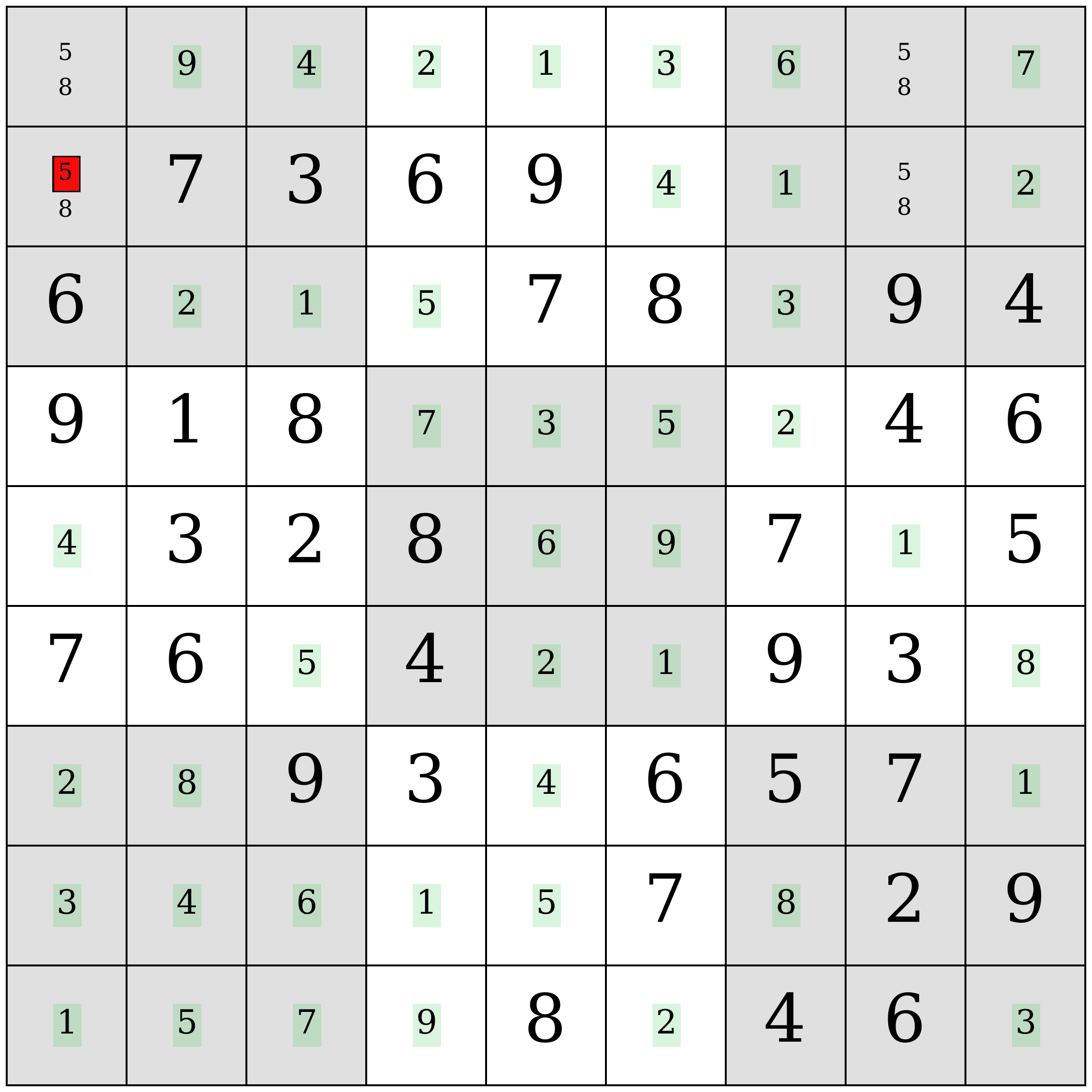}}
  \frame{\includegraphics[scale=0.17]{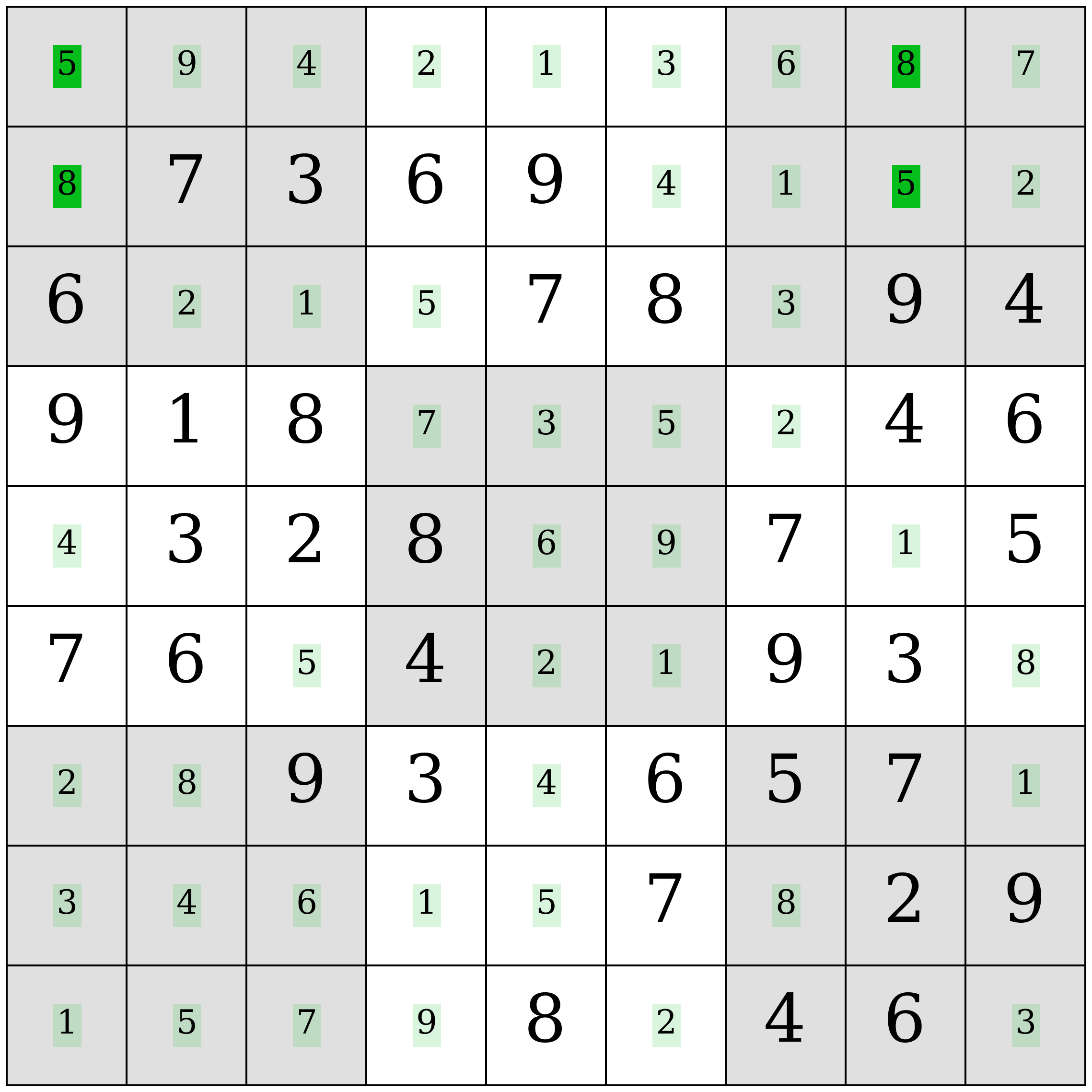}}
\caption{Visualizations of the stages while solving a Sudoku puzzle}
  \label{fig:sudokuframes}
\end{figure}

To illustrate the process described above,
we use the Sudoku puzzle from \clingraph's examples folder.\footnote{\url{https://github.com/potassco/clingraph/tree/master/examples/propagator/sudoku}}
In this encoding,
we use
predicate \lstinline[mathescape]{sudoku($x$,$y$,$v$)} to represent a
cell with coordinates ($x$,$y$) in a  $9\times 9$ grid
with an assigned digit $v$ from 1 to 9.
A cell can have an initial value defined in the instance
by predicate \lstinline[mathescape]{initial($x$,$y$,$v$)}
or it can be empty if no such predicate appears.
Then,
the problem encoding and instance are handed to
\clingo's solving process which is observed by our propagator.
Partial assignments accumulated by the propagator are passed to the visualization encoding,
which is shown in Table~\ref{lst:vizsudoku}.
Additionally, Figure~\ref{fig:sudokuframes} depicts the resulting animation's
key frames visualizing
the partial assignments reached during solving.
\begin{table}[!ht]\begin{minipage}{\linewidth}
\begin{lstlisting}[language=clingo,escapeinside={\#\#\%}{\#},basicstyle=\scriptsize\ttfamily,firstnumber=38,linerange={38-39}]
attr(node, pos(X,Y), fontsize,40 ) :- _true(initial(X,Y,V)).##%\label{lst:vizsudoku:initial:font}#
attr(node, pos(X,Y), label, V ) :- _true(initial(X,Y,V)).##%\label{lst:vizsudoku:initial:label}#
\end{lstlisting}
\begin{lstlisting}[language=clingo,escapeinside={\#\#\%}{\#},basicstyle=\scriptsize\ttfamily,firstnumber=42,linerange={42-48}]
attr(node, pos(X,Y), fontsize, 20) :- _true(sudoku(X,Y,V)), not _true(initial(X,Y,_)).##%\label{lst:vizsudoku:onevalue:font}#
attr(node, pos(X,Y), label, @concat("<<table BORDER='0'>",##%\label{lst:vizsudoku:onevalue:labelstart}#
  "<tr><td BGCOLOR='{{color[",V,"]}}{{opacity[",V,"]}}'",
         " BORDER='{{border[",V,"]}}'>",
         "{{value[",V,"]}}",
  "</td></tr></table>>")) :-
      _true(sudoku(X,Y,V)), not _true(initial(X,Y,_)).##%\label{lst:vizsudoku:onevalue:labelend}#
\end{lstlisting}
\begin{lstlisting}[language=clingo,escapeinside={\#\#\%}{\#},basicstyle=\scriptsize\ttfamily,firstnumber=52,linerange={52-69}]
attr(node, pos(X,Y), label, @concat("<<table BORDER='0'>",##%\label{lst:vizsudoku:multvalue:labelstart}#
  "<tr>",
    "<td BGCOLOR='{{color[1]}}{{opacity[1]}}' BORDER='{{border[1]}}'>{{value[1]}}</td>",
    "<td BGCOLOR='{{color[2]}}{{opacity[2]}}' BORDER='{{border[2]}}'>{{value[2]}}</td>",
    "<td BGCOLOR='{{color[3]}}{{opacity[3]}}' BORDER='{{border[3]}}'>{{value[3]}}</td>",
  "</tr>",
  "<tr>",
    "<td BGCOLOR='{{color[4]}}{{opacity[4]}}' BORDER='{{border[4]}}'>{{value[4]}}</td>",
    "<td BGCOLOR='{{color[5]}}{{opacity[5]}}' BORDER='{{border[5]}}'>{{value[5]}}</td>",
    "<td BGCOLOR='{{color[6]}}{{opacity[6]}}' BORDER='{{border[6]}}'>{{value[6]}}</td>",
  "</tr>",
  "<tr>",
    "<td BGCOLOR='{{color[7]}}{{opacity[7]}}' BORDER='{{border[7]}}'>{{value[7]}}</td>",
    "<td BGCOLOR='{{color[8]}}{{opacity[8]}}' BORDER='{{border[8]}}'>{{value[8]}}</td>",
    "<td BGCOLOR='{{color[9]}}{{opacity[9]}}' BORDER='{{border[9]}}'>{{value[9]}}</td>",
  "</tr>",
  "</table>>")) :- 
      _true(pos(X,Y)), not _true(initial(X,Y,_)), not _true(sudoku(X,Y,_)).##%\label{lst:vizsudoku:multvalue:labelend}#
\end{lstlisting}
\begin{lstlisting}[language=clingo,escapeinside={\#\#\%}{\#},basicstyle=\scriptsize\ttfamily,firstnumber=74,linerange={74-77}]
attr(node, pos(X,Y), (label,opacity,V), 25  ) :- not _change(sudoku(X,Y,V)),##%\label{lst:vizsudoku:opacity:start}#
       not _decide(sudoku(X,Y,V),_), _true(sudoku(X,Y,V)).##%\label{lst:vizsudoku:opacity:end}#
attr(node, pos(X,Y), (label,opacity,V), "00") :- not _true(sudoku(X,Y,V)),##%\label{lst:vizsudoku:transp:start}#
       not _decide(sudoku(X,Y,V),neg), _true(pos(X,Y)), value(V).##%\label{lst:vizsudoku:transp:end}#
\end{lstlisting}
\begin{lstlisting}[language=clingo,escapeinside={\#\#\%}{\#},basicstyle=\scriptsize\ttfamily,firstnumber=80,linerange={80-82}]
attr(node, pos(X,Y), (label,border,V), 1) :- _decide(sudoku(X,Y,V),_).##%\label{lst:vizsudoku:border}#
attr(node, pos(X,Y), (label,border,V), 0) :- not _decide(sudoku(X,Y,V),_),##%\label{lst:vizsudoku:noborder:start}#
      _true(pos(X,Y)), value(V).##%\label{lst:vizsudoku:noborder:end}#
\end{lstlisting}
\begin{lstlisting}[language=clingo,escapeinside={\#\#\%}{\#},basicstyle=\scriptsize\ttfamily,firstnumber=85,linerange={85-89}]
attr(node, pos(X,Y), (label,color,V), white) :- not _true(sudoku(X,Y,V)),##%\label{lst:vizsudoku:whitecol:start}#
      not _decide(sudoku(X,Y,V),_), _true(pos(X,Y)), value(V).##%\label{lst:vizsudoku:whitecol:end}#
attr(node, pos(X,Y), (label,color,V), green) :- _true(sudoku(X,Y,V)).##%\label{lst:vizsudoku:truecolor}#
attr(node, pos(X,Y), (label,color,V), red  ) :- _decide(sudoku(X,Y,V),neg).##%\label{lst:vizsudoku:decide:neg}#
attr(node, pos(X,Y), (label,color,V), green) :- _decide(sudoku(X,Y,V),pos).##%\label{lst:vizsudoku:decide:pos}#
\end{lstlisting}
\begin{lstlisting}[language=clingo,escapeinside={\#\#\%}{\#},basicstyle=\scriptsize\ttfamily,firstnumber=92,linerange={92-94}]
attr(node, pos(X,Y), (label,value,V), V ) :- _true(sudoku(X,Y,V)).##%\label{lst:vizsudoku:trueval}#
attr(node, pos(X,Y), (label,value,V), V ) :- _undefined(sudoku(X,Y,V)).##%\label{lst:vizsudoku:undefval}#
attr(node, pos(X,Y), (label,value,V), "") :- _false(sudoku(X,Y,V)).##%\label{lst:vizsudoku:falseval}#
\end{lstlisting}
\end{minipage}
  \caption{Selected lines from the encoding visualizing Sudoku solving (\texttt{viz-sudoku-solving.lp})}
  \label{lst:vizsudoku}
\end{table}

Let us now examine how the frames from Figure~\ref{fig:sudokuframes} are constructed.
Each cell with an initial value is visualized by
setting the corresponding digit as the label of its node
(rule in Line~\ref{lst:vizsudoku:initial:label} from Table~\ref{lst:vizsudoku}) and
using a relatively larger font size (rule in Line~\ref{lst:vizsudoku:initial:font}).
These rules have the reified literal \lstinline{_true(initial(X,Y,V))} in the body to represent cells with initial values.
Notice that facts appearing in the problem input, such as \lstinline{initial(X,Y,V)},
will always have their truth value set to true.
For each node of an empty cell,
we construct an HTML-like label
that allows us to use rich visual elements
like tables with different borders and background colors.
In order to ease constructing long HTML-like labels
we rely on template strings (see Section~\ref{sec:template}).
Let us first cover empty cells that must be filled with one specific digit.
The HTML-like label for such a cell represents a table
having only one slot for the respective digit.
The rule in Lines~\ref{lst:vizsudoku:onevalue:labelstart} to \ref{lst:vizsudoku:onevalue:labelend}
generates such a label as a template string
by concatenating the constituent strings
using the \lstinline{concat} external function provided by \clingraph.
Note that the rule body designates
an initially empty cell (captured by the body literal \lstinline{not _true(initial(X,Y,_))})
that must be filled with a specific digit (\lstinline{_true(sudoku(X,Y,V))}).
We set the font size of these cells via the rule in Line~\ref{lst:vizsudoku:onevalue:font}.
An example of this can be found on the third cell in the topmost row of the top leftmost graph
in Figure~\ref{fig:sudokuframes},
where this initially empty cell is now filled with digit~4 with dark green background.
For adding style to such cells,
the label template uses variables to represent the color, opacity, border and value.
The values for these variables are obtained through different rules
that generate atoms over \lstinline{attr/4}.
For this specific cell,
the RGB code of dark green as the value of variable \lstinline|color[V]| is set via
the rule in Line~\ref{lst:vizsudoku:truecolor}.
Furthermore, the rule in Line~\ref{lst:vizsudoku:noborder:start} to \ref{lst:vizsudoku:noborder:end}
assigns value \lstinline{0} to variable \lstinline|border[V]| to avoid setting borders
and the rule in Line~\ref{lst:vizsudoku:trueval} assigns \lstinline{4} to \lstinline|value[V]|.
We also add opacity to the background color codes to highlight
changes in the current partial assignment from the ones propagated in previous assignments
by reducing the opacity of older ones.
In this specific cell,
since variable \lstinline|opacity[V]| gets empty string as a default value
due to absence of any rule generating a specific value for the variable,
the opacity of the background color is not modified.
The same cell but in the following graph, has a light green background color,
which designates that \clingo\ filled it in an earlier propagation step.
In order to generate the light green color,
in that step variable \lstinline|opacity[V]| gets its value \lstinline{25}
via the rule in Line~\ref{lst:vizsudoku:opacity:start} to \ref{lst:vizsudoku:opacity:end}.
Note that its body literal \lstinline{not _change(sudoku(X,Y,V))} captures
the respective cell is not filled in the current partial assignment.
These rules generating variable values are also used for the Sudoku cells that have multiple options.
We describe such cells below.

The remaining type of cells are those initially empty cells in which more than one digit may appear in a partial assignment.
Their HTML-like label represents a 3 $\times$ 3 table
allowing a slot for each digit from 1 to 9.
Our aim is to visualize digits that can possibly appear in such a cell
in this tabular form.
For instance,
the top leftmost graph
shows that either 2, 5 or 8 can be placed in the first cell.
The rule in Lines~\ref{lst:vizsudoku:multvalue:labelstart} to \ref{lst:vizsudoku:multvalue:labelend}
constructs the label as a template string representing the 3 $\times$ 3 table.
The body literal \lstinline{not _true(sudoku(X,Y,_))} captures these initially empty
cells with multiple options.
Consider \clingo\ has reasoned that digit $d$ from 1 to 9 cannot appear in an empty cell $(x,y)$ in
a partial assignment,
which is represented by the reified fact \lstinline[mathescape]{_false(sudoku($x$,$y$,$d$))}
generated by the propagator (e.g., \lstinline{_false(sudoku(1,9,1))} for the cell mentioned above).
We do not show $d$ in its respective slot.
To this end, template variable \lstinline|value[d]| is assigned to empty string
via the rule in Line~\ref{lst:vizsudoku:falseval}.
Additionally, its transparent background color is controlled via
rules in Line~\ref{lst:vizsudoku:whitecol:start} to \ref{lst:vizsudoku:whitecol:end}
and in Line~\ref{lst:vizsudoku:transp:start} to \ref{lst:vizsudoku:transp:end}
by setting \lstinline{white} to variable \lstinline|color[d]| and
\lstinline{"00"} to variable \lstinline|opacity[d]|, respectively.
Also consider \clingo\ may be undecided on whether digit $d$ is the value of the empty cell or not
(e.g., digits 2, 5 and 8 for the cell mentioned above).
This is reflected by the fact \lstinline[mathescape]{_undefined(sudoku($x$,$y$,$d$))}
generated by the propagator for a partial assignment.
We show $d$ in its respective slot by setting
variable \lstinline|value[d]| to $d$ this time via the rule in Line~\ref{lst:vizsudoku:undefval}.
Its transparent background is set via the same rules in
Line~\ref{lst:vizsudoku:whitecol:start} to \ref{lst:vizsudoku:whitecol:end}
and Line~\ref{lst:vizsudoku:transp:start} to \ref{lst:vizsudoku:transp:end}.
We can also visualize whenever the propagation during solving reaches a fixpoint,
and \clingo\ may decide on a truth value of an undefined atom to continue search.
For instance, the second graph in the first row of Figure~\ref{fig:sudokuframes} shows
such a decision point as digit 5 in red background with a border
where \clingo\ selects the atom \lstinline{sudoku(4,2,5)} to be false.
Its background color and border are set via rules in
Line~\ref{lst:vizsudoku:decide:neg} and Line~\ref{lst:vizsudoku:border}, respectively.
Whenever \clingo\ selects an atom to be true at a decision point,
we visualize it as green (rule in Line~\ref{lst:vizsudoku:decide:pos}).

Ultimately, our animation allows us to analyze different aspects of the solving process of the Sudoku.
For instance, the first graph illustrates that during the initial propagation
\clingo\ already fills many cells with digits (those having digits with green background)
and constrains the remaining empty cells that only possible digits are shown.
This can be an indicator of how simple the Sudoku instance is.
Finally, when we reach the last graph (bottom rightmost)
passing through various stages of solving in order,
we get an answer set representing a solution of the puzzle instance.

 \subsection{Visualizing the program structure}\label{subses:vizprogram}

So far, we have visualized the solving process, input and/or result of a program.
However, we may also visualize information about the program itself.
In this section, we concentrate on the abstract syntax tree (AST) of a program,
as it is accessible via the \lstinline{clingo.ast} module of \clingo's Python API.
Visualizing a program's AST eases the understanding of its internal structure.
This is of particular interest when dealing with non-ground programs.
To this end, we follow a three-stage process.
First, we translate the AST into an intermediate fact format using a simple Python script called \lstinline{reify_ast.py}.
Then, we employ a visualization encoding to convert these facts into \clingraph's format.
Finally, we run \clingraph\ to render the graph.
All examples can be found in \clingraph's repository.\footnote{\url{https://github.com/potassco/clingraph/tree/master/examples/ast}}

Our intermediate fact format uses two predicates:
\lstinline{ast_node/3} and \lstinline{ast_edge/3} represent the nodes and edges in an AST.
These predicates are meant to be semantic triples,
linking a subject, in our case a node or edge, via a key to a value.
This is inspired by the representation of graph data, as used in the Resource Description Framework\footnote{\url{https://www.w3.org/TR/rdf11-concepts/}}.
To illustrate how ASTs are stored in terms of triples,
let us look at the translation of the program in Listing~\ref{lst:color}.
It is partially depicted in Listing~\ref{lst:color_ast} and may be obtained by running \lstinline{reify_ast.py color.lp}.
\begin{lstlisting}[language=clingos,basicstyle=\scriptsize\ttfamily,label={lst:color_ast},caption={Partial translation of the program in Listing~\ref{lst:color} into our intermediate fact format}]
ast_node(875,type,"AST").%%#(\label{lst:color_ast:type}#)
ast_node(875,variant,"Rule").%%#(\label{lst:color_ast:variant}#)
ast_node(875,value,"#false :- edge(N,M), assign(N,C), assign(M,C).").%%#(\label{lst:color_ast:value}#)
ast_edge((875,885),key,"head").%%#(\label{lst:color_ast:keyA}#)
ast_edge((875,898),key,"body").%%#(\label{lst:color_ast:keyO}#)
\end{lstlisting}

Nodes are represented by unique integers which are arbitrarily chosen by our script.
Edges are identified using the integers of the adjacent nodes.
Line~\ref{lst:color_ast:type} tells us that a node~875 exists and that it stands for an instance of the class
\lstinline{clingo.ast.AST} in the Python API of \clingo.
The other possible types are \lstinline{ASTSequence}, \lstinline{Location}, \lstinline{Position}, \lstinline{Symbol},
\lstinline{int}, \lstinline{str} and \lstinline{None}.
They cover the respective classes in \lstinline{clingo.ast} and the necessary basic types in Python.
If the type of a node is \lstinline{AST}, declaring a variant as in Line~\ref{lst:color_ast:variant} is mandatory.
The variant reflects the \lstinline{clingo.ast.ASTType} of each instance of \lstinline{clingo.ast.AST}.
The possible variants include \lstinline{Rule}, \lstinline{Variable}, \lstinline{SymbolicAtom} and many more.
Line~\ref{lst:color_ast:value} assigns a value to node~875,
in this case a string representing the rule represented by node~875.
Line~\ref{lst:color_ast:keyA} and \ref{lst:color_ast:keyO} reflect two outgoing edges of node~875, named \lstinline{head} and \lstinline{body}.
As the names suggest, these edges point to nodes capturing the head and body of the rule.

\begin{figure}[ht!]
    \centering
    \includegraphics[width=\textwidth]{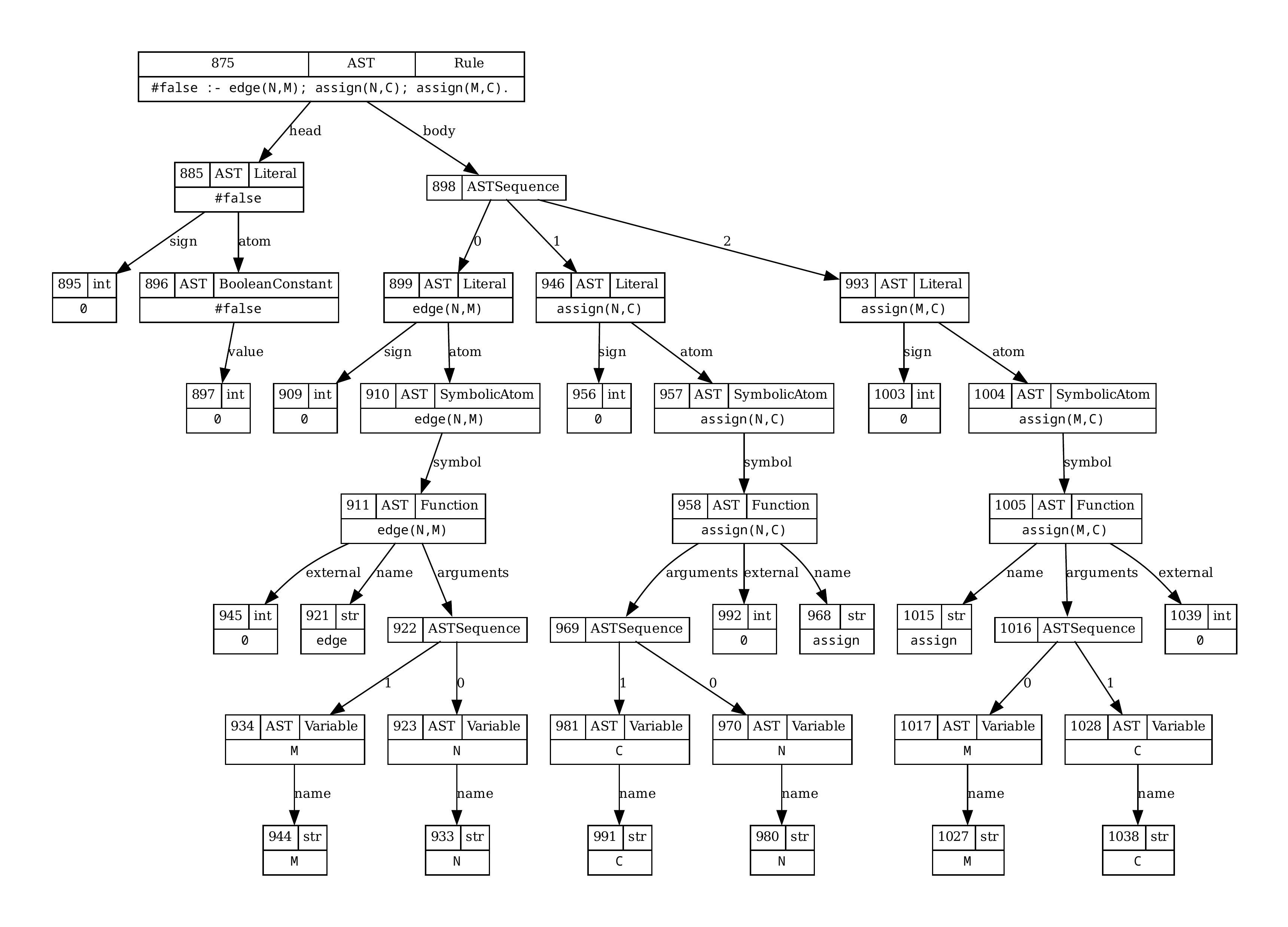}
    \caption{A partial visualization of the AST of the program in Listing~\ref{lst:color}.}
    \label{fig:ast}
\end{figure}

\begin{lstlisting}[language=clingos,basicstyle=\scriptsize\ttfamily,label={lst:viz_ast},caption={Selected lines from the encoding visualizing the AST (\lstinline{viz-ast.lp})},firstline=4,lastline=47]
ast_show(edge, I) :- ast_edge(I, _, _), I = (I1, _), ast_show(node, I1).%%#(\label{lst:viz_ast:show_hide:begin}#)
ast_show(node, I) :- ast_show(edge, (_, I)).

ast_hide(node, I) :- ast_hide(edge, (_, I)).
ast_hide(edge, I) :- ast_edge(I, _, _), I = (I1, _), ast_hide(node, I1).%%#(\label{lst:viz_ast:show_hide:end}#)

node(I) :- ast_node(I, _, _), ast_show(node, I), not ast_hide(node, I).
edge(I) :- ast_edge(I, _, _), ast_show(edge, I), not ast_hide(edge, I).

attr(node, I, label, @concat(%%#(\label{lst:viz_ast:template:begin}#)
    "<<table border='0' cellborder='1' cellspacing='0' cellpadding='3'>",
    "<tr>",
        "<td>{{id}}</td>",
        "<td>{{type}}</td>",
        "{% if variant %}<td>{{variant}}</td>{% endif %}",
    "</tr>",
    "{% if value %}",
        "<tr><td colspan='{{colspan}}'>",
            "<font face='monospace'> {{value}} </font>",
        "</td></tr>",
    "{% endif %}",
    "</table>>"
)) :- node(I).%%#(\label{lst:viz_ast:template:end}#)

attr(node, I, (label, id), I)
    :- node(I).
attr(node, I, (label, type), T)
    :- node(I), ast_node(I, type, T).
attr(node, I, (label, variant), V)
    :- node(I), ast_node(I, variant, V).
attr(node, I, (label, colspan), 2)
    :- node(I), ast_node(I, value, _), not ast_node(I, variant, _).
attr(node, I, (label, colspan), 3)
    :- node(I), ast_node(I, value, _), ast_node(I, variant, _).
attr(node, I, (label, value), @html_escape(V))
    :- node(I), ast_node(I, value, V).

attr(edge, I, label, L) :- edge(I), ast_edge(I, key, L).

attr(graph_nodes, default, fontsize, 10).
attr(graph_nodes, default, shape, plain).

attr(graph_edges, default, fontsize, 10).
\end{lstlisting}
In order to translate the output of our script into the input format of \clingraph, we employ a visualization encoding,
assembling an HTML-like label including all the data stored in the semantic triples.
Its main component is the template (see Section~\ref{sec:template}) shown in Listing~\ref{lst:viz_ast} from Line~\ref{lst:viz_ast:template:begin} to \ref{lst:viz_ast:template:end}.
Given that even small programs have large syntax trees, our encoding provides functionalities (Line~\ref{lst:viz_ast:show_hide:begin} to \ref{lst:viz_ast:show_hide:end}) to show and hide subtrees.
For instance,
using \lstinline{ast_show(node, 875).} guarantees that only the subtree of node~875 is shown,
while the rule
\begin{lstlisting}[numbers=none]
ast_hide(edge, I) :- ast_edge(I, _, _), I = (_, I2),
                     ast_node(I2, type, "Location").
\end{lstlisting}
allows us to hide any subtree that is rooted at a node of type \lstinline{Location}.
Calling
\begin{lstlisting}[numbers=none,basicstyle=\ttfamily\small]
reify_ast.py color.lp | \
clingraph --viz-encoding=viz-ast.lp --type=digraph \
          --out=render --format=pdf
\end{lstlisting}
with the above line in the visualization encoding instructs \clingraph\ to render the graph shown in Figure~\ref{fig:ast}.

Showing the abstract syntax tree is by far not the only option to visualize a program's structure.
In principle, \clingraph\ may render any structured knowledge about the program provided that
a reification format, a tool generating it, and a visualization encoding exists.
To this end, our case study may serve as a blueprint for future ideas.

  \section{Formatting attributes with templates}
\label{sec:template}

Generating complex string values for attributes can become quite cumbersome,
especially when dealing with HTML-like labels.\footnote{\url{https://graphviz.org/doc/info/shapes.html#html}}
This type of \graphviz\ labels are formed by an HTML string delimited by \lstinline{<...>} which gives a lot of flexibility for formatting the text and generating tables.
We simplify the generation of such strings by using the template engine Jinja.\footnote{\url{https://jinja.palletsprojects.com/en/3.1.x/templates}}
Attribute values can then be seen as Jinja templates,
which are rendered using the variables provided by atoms of the form \lstinline[mathescape]{attr($t$,$id$,($n$,$v$),$x$)}.
In these atoms,
the third argument is a pair indicating that variable $\mathit{v}$ has value $\mathit{x}$ when rendering the template of attribute $\mathit{n}$.
Furthermore, we can encapsulate values in dictionary variables by using triples instead,
where $(n,v,k)$ indicates that variable $\mathit{v}$ is a dictionary with entry $\{k:x\}$.
When no template is provided, the values of all the variables are concatenated.

\begin{figure}[ht!]
    \centering
    \includegraphics[scale=0.25]{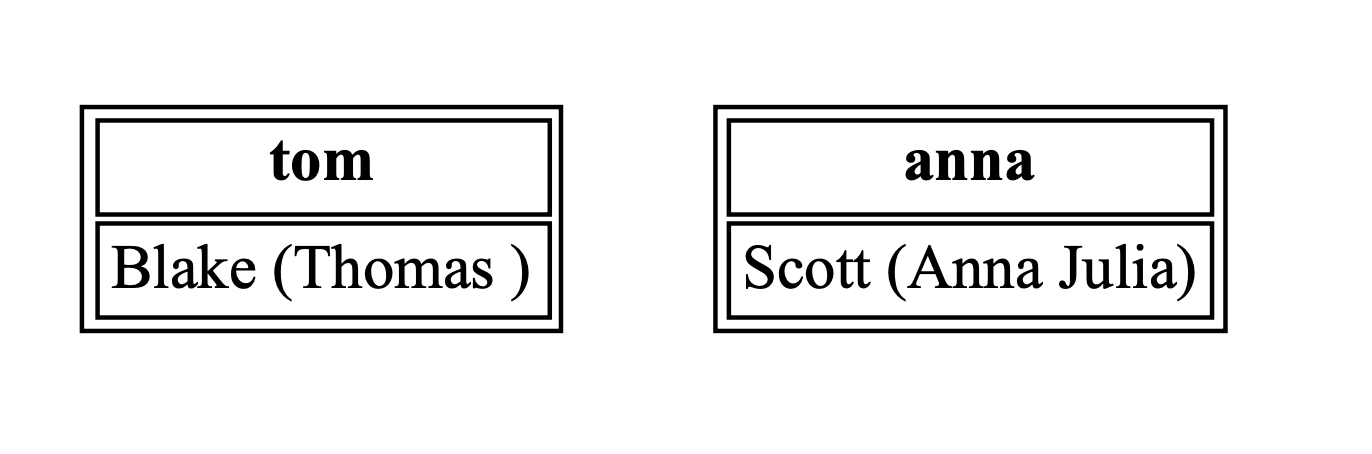}
    \caption{Example of HTML-like labels using attribute templates.}
    \label{fig:template}
\end{figure}

For illustration, we visualize the data of people, defined by predicates
\lstinline{person/1}, \lstinline{name/1}, \lstinline{middlename/1} and \lstinline{lastname/1}
(see Listing~\ref{lst:template-instance} for an example).
We visualize the data using HTML-like labels to generate the tables in Listing~\ref{lst:template}.
\begin{lstlisting}[float,language=clingo,label={lst:template},basicstyle=\scriptsize\ttfamily,caption={Visualization encoding to exemplify the generation of strings via templates (\texttt{template.lp}).}]
node(N) :- person(N).                                                                   %%#(\label{lst:template:node}#)
attr(node,N,shape,none) :- person(N).                                                   %%#(\label{lst:template:shape}#)
attr(node,N,label, @concat( %%#(\label{lst:template:label}#)
  "<<table>",                                                             %%#(\label{lst:template:label:one}#)
    "<tr><td><b>{{id}}</b></td></tr>",
    "<tr><td>{{lastname}} ({{name['first']}} {{name['middle']}})</td></tr>",
  "</table>>")) :- person(N).                                             %%#(\label{lst:template:label:two}#)
attr(node,N,(label,id),N) :- person(N).                                              %%#(\label{lst:template:id}#)
attr(node,N,(label,name,first),Name) :- name(N,Name).                                   %%#(\label{lst:template:first}#)
attr(node,N,(label,name,middle),Name) :- middlename(N,Name).                            %%#(\label{lst:template:middle}#)
attr(node,N,(label,lastname),Lastname) :- lastname(N,Lastname).                         %%#(\label{lst:template:last}#)
\end{lstlisting}
Line~\ref{lst:template:node} generates a node for each person and
Line~\ref{lst:template:shape} removes the shape of the node (no shape is needed since the label is a table).
Line~\ref{lst:template:label} to~\ref{lst:template:label:two} define the template for the label using the HTML tags
\lstinline{<table>}, \lstinline{<tr>}, \lstinline{<td>} and  \lstinline{<b>}
to construct a table, row, cell, and boldface text, respectively.
Variables are enclosed in double braces \lstinline|{{...}}| and
corresponding values are substituted by Jinja for these variables.
The rules in Line~\ref{lst:template:id} to~\ref{lst:template:last}, for instance,
generate atoms of \lstinline{attr/4} to populate values for the template variables.
Line~\ref{lst:template:id} and Line~\ref{lst:template:last} use a pair to assign $N$
to the variable $\mathit{id}$ and the last name to $\mathit{lastname}$.
Unlike,
Line~\ref{lst:template:first} and~\ref{lst:template:middle} use a triple,
making the variable $\mathit{name}$ a dictionary with the keys $\mathit{first}$ and $\mathit{middle}$,
which is accessed in the template as $\mathit{name['first']}$ and $\mathit{name['middle']}$, respectively.
The output of this encoding together with the instance defined in Listing~\ref{lst:template-instance}
is shown in Figure~\ref{fig:template}.
\begin{lstlisting}[float,language=clingo,label={lst:template-instance},basicstyle=\scriptsize\ttfamily,caption={Instance for the template example (\texttt{people.lp}).}]
person(anna). 
name(anna,"Anna"). middlename(anna,"Julia"). lastname(anna,"Scott").
person(tom). 
name(tom,"Thomas"). lastname(tom,"Blake").
\end{lstlisting}
It produced by means of the following instruction:
\begin{lstlisting}[basicstyle=\small\ttfamily,numbers=none,xleftmargin=\parindent]
clingraph people.lp --viz-encoding=template.lp --out=render
\end{lstlisting}

Jinja's syntax of templates also includes statements like conditionals, loops, and several operations.
We refer the interested reader to our github repository for more complex examples of \clingraph\ using such features.\footnote{\url{https://github.com/potassco/clingraph/tree/master/examples/office}}

 \newcommand{\kara}{\sysfont{kara}}
\newcommand{\lonsdaleite}{\sysfont{lonsdaleite}}
\newcommand{\idpdraw}{\sysfont{idpdraw}}
\newcommand{\nomore}{\sysfont{nomore}}
\newcommand{\sealion}{\sysfont{sealion}}
\section{Related work}
\label{sec:related:work}
Many aspects of \clingraph\ are inspired by previous systems described in the literature.
The basic goal---to visualize answer sets by mapping special atoms to graphic elements---traces back to \aspviz~\cite{clvobrpa08a},
a command-line application written in Java using the Standard Widget Toolkit (SWT) for rendering.
It is capable of rendering two-dimensional graphics with absolute coordinates but does neither allow relative positioning nor graph structures.
These features were introduced by \kara~\cite{kloeputo11a}, a plugin written for the SeaLion IDE.
The alternative of using \graphviz\ as a backend was first mentioned by the authors of \aspviz, and followed up with a
rather basic implementation in \lonsdaleite\footnote{\url{https://github.com/rndmcnlly/Lonsdaleite}}.
Another visualizer for answer sets is \idpdraw\footnote{\url{https://dtai.cs.kuleuven.be/krr/files/bib/manuals/IDPDraw-manual.pdf}}, although it seems to be discontinued.

The idea of visualizing the solving process was first explored for the \nomore\ system~\cite{bolisc04a} which uses a graph-oriented computational model.
For \dlv, there exists a graphical tool for developing and testing logic programs~\cite{peritecive07a} as well as a visual tracer~\cite{calerive09a}.
In the realms of \clingo, visualizing the solving process has been explored using a tweaked version of \clasp~\cite{koesch13a}.

Our system not only integrates ideas from the literature and makes them available for modern ASP systems, but also has some features that have---to the best of our knowledge---never been implemented before.
There is a powerful API which makes it easy to include \clingraph\ in custom projects,
a multitude of different output formats including \LaTeX\ and animated GIF,
and the capacity of integrating a propagator for visualizing the solving process of \clingo.

 \section{Discussion}\label{sec:discussion}

\Clingraph\ provides essentially an ASP-based front-end to the graph visualization software \graphviz.
In doing so, it takes up the early approach of \aspviz~\cite{clvobrpa08a} and extends it in the context of modern ASP technology.
The advantage of \clingraph\ is that one does not have to resort to foreign programming languages for visualization
but rather remains within the realm of ASP.
This provides users with an easy interface among logic programs and/or answer sets and their visualization.
Moreover, \clingraph\ offers a \python\ API that extends this ease of interfacing to \clingo's API,
and in turn to connect and monitor various aspects of the solving process.
The fact-based interface of \clingraph\ makes it readily applicable to any ASP system.
For more advanced features,
like json output and API functionality,
\clingraph\ depends on \clingo.
\Clingraph\ is open source software and freely available at \url{https://github.com/potassco/clingraph}.

\paragraph{Acknowledgments}
This work was supported by DFG grants SCHA 550/11 and~15 as well as BMBF project ISCO with support code KK5291302GR1.

\nocite{hasascst22a}

\end{document}